\documentclass[letterpaper, 10 pt, conference]{ieeeconf} 
\long\def\/*#1*/{}
\IEEEoverridecommandlockouts                              % This command is only needed if 
\usepackage{color}
\usepackage{soul}
\usepackage{amsmath}
\usepackage[pdftex]{graphicx}
\usepackage{listings}
\usepackage{gensymb}
\usepackage{tikz}
\usepackage{subfig}
\usepackage{authblk}
\usepackage[pdftex]{graphicx}
\usepackage{todonotes,amsmath,amsbsy,amssymb,latexsym,algorithm,algorithmic,url,verbatim}
\newcommand{\Fig}[1]{Fig.~\ref{fig:#1}}
\begin{document}
\title{Modeling Basic Aspects of Cyber-Physical Systems, Part II}   
%\preprintfooter{Under review} % 'preprint' option specified.

\newcommand{\hide}[1]{}

\author[1]{Yingfu Zeng\thanks{This paper is a followup to a paper
    presented at the DSLRob 2012 Workshop with the same main
    title~\cite{taha2013modeling}. This second part focuses on
    modeling rigid body dynamics.}\thanks{This work was supported
    by the US NSF CPS award 1136099, Swedish KK-Foundation CERES and
    CAISR Centres, and the Swedish SSF NG-Test Project.}}
\author[1]{Chad Rose} \author[1]{Paul Brauner} \author[1,2]{Walid
  Taha} \author[2]{Jawad Masood} \author[2]{Roland Philippsen}
\author[1]{\\Marcia O'Malley} \author[1,2]{Robert Cartwright}
\affil[1]{Rice University} \affil[2]{Halmstad University}

\hide{
\author{Yingfu Zeng$^\ddagger$,
            Chad Rose$^\ddagger$,
            Jawad Masood $^\diamond$,
            Roland Philippsen $^\diamond$,
            Paul Brauner $^\ddagger$,
            Walid Taha $^\diamond$$^\ddagger$,
            Marcia O'Malley $^\ddagger$,
            Robert Cartwright $^\ddagger$$^\diamond$}
           {$^\ddagger$Rice University and $^\diamond$Halmstad University}
           {yz39@rice.edu}
\hide{
\authorinfo{\vspace{-20}Yingfu Zeng$^\star$}
           {\vspace{-5}}
           {yz39@rice.edu}
\authorinfo{\vspace{-20}Chad Rose$^\star$}
           {\vspace{-5}}
           {chad.g.rose.rice@gmail.com}
\authorinfo{\vspace{-20}Jawad Masood$^\diamond$}
           {\vspace{-5}}
	{jawwadmasood@gmail.com}
\authorinfo{\vspace{-25} Roland Philippsen$^\diamond$}
           {\vspace{-5}}
           {Roland.Philippsen@hh.se}
\authorinfo{\vspace{-25}Paul Brauner$^\star$}
	{\vspace{-5}Current Affiliation: Google Inc}
	{polux@google.com}
\authorinfo{\vspace{-25}Walid Taha$^\diamond$$^\star$}
           {\vspace{-5}}
           {Walid.Taha@hh.se}
\authorinfo{\vspace{-30}Marcia O'Malley$^\star$}
	{\vspace{-5}}
	{omalleym@rice.edu}
\authorinfo{\vspace{-30}Robert Cartwright$^\star$}
	{\vspace{-5}}
	{cork@rice.edu}
}
}

\maketitle \begin{abstract} We continue to consider the question of
  what language features are needed to effectively model
  cyber-physical systems (CPS).  In previous work, we proposed using a
  core language as a way to study this question, and showed how
  several basic aspects of CPS can be modeled clearly in a language
  with a small set of constructs.  This paper reports on the result of
  our analysis of two, more complex, case studies from the domain of
  rigid body dynamics.  The first one, a quadcopter, illustrates that
  previously proposed core language can support larger, more
  interesting systems than previously shown.  The second one, a serial
  robot, provides a concrete example of why we should add language
  support for static partial derivatives, namely that it would
  significantly improve the way models of rigid body dynamics can be
  expressed.
\end{abstract}

\section{Introduction} 

The increasing computational power embedded in everyday products
promises to revolutionize the way we live.  At the same time, the
tight coupling between computational and physical mechanisms, often
described as cyber-physical systems (CPSs), poses a challenge for the
traditional product development cycles, particularly in physical
testing.  For example, car manufacturers are concerned about the
amount of physical testing necessary to assure the safety of
autonomous vehicles.  Physical testing has been used to assess the
qualities of new products for many years.  One of its key ingredients
is devising a collection of specific test scenarios.  But the presence
of even simple computational components can make it difficult to
identify enough test scenarios to exercise more than a minute fraction
of the possible behaviors of the system.  In addition, physical
testing is very expensive because it only detects flaws at the end of
the product development process after physical prototypes have been
constructed.  These realities are spurring the CPS developers to
rethink traditional methods and processes for developing and testing
new products.

Computer simulations~\cite{carloni-2006} performing virtual
experiments~\cite{bruneau-2012} can be used to reduce the cost of
physical tests.  Virtual testing can be used to quickly eliminate
obviously bad designs.  It can also help build confidence that a new
design can pass test scenarios developed by an independent
party~\cite{jensen-2011}.  However, creating a framework for
conducting virtual experiments requires a concerted, interdisciplinary
effort to address a wide range of challenges, including: 1) educating
designers in the cyber-physical aspects of the products they will
develop, both in terms of how these aspects are modeled, and what
types of system-level behaviors are generated; and 2) developing
expressive, efficient, and robust modeling and simulation tools to
support the innovation process.  At each stage in the design process,
the underlying models should be easy to understand and analyze.
Moreover, it should be easy to deduce the mathematical relationship
between the models used in successive stages.

Both challenges can be addressed by better language-based technologies
for modeling and simulation. An effective model should have clear and
unambiguous semantics that can readily be simulated, showing how the
model behaves in concrete scenarios.  Engineering methods centered
around a notion of executable or effective models can have profound
positive impacts on the pace of advancement of knowledge and
engineering practice.

\subsection{The Accessibility Challenge}
Designing a future smart vehicle or home requires expertise from a
number of different disciplines.  Even when we can assemble the
necessary team of experts, they often lack a common language for
discussing key issues -- treated differently across disciplines --
that arise in CPS design.

A critical step towards addressing this {\em accessibility challenge}
is to discover (and not ``invent'') a {\em lingua franca} (or ``common
language'') that can break down artificial linguistic barriers between
various scientific and engineering disciplines. Part of such a common
language will be a natural language that includes a collection of
technical terms from the domain of CPS, and that enables experts to
efficiently express common model constructions; part will be
executable (meaning computationally effective) modeling formalisms
made up from a subset of the mathematical notation {\em already in
  use}.  Language research can be particularly helpful in discovering
the latter part \cite{zhu2010mathematical}.  Examples of such tools
exist, but are usually not available as widely-used, multi-purpose
tools.  This is the case for languages such as Fortress
\cite{fortress} and equational languages \cite{Modelica,Simscape}, as
well as specialized, physics-specific or even multi-physics tools.
When they are available as widely-used, multi-purpose tools, they
generally cannot be viewed as executable formalisms.  This is the case
for most symbolic algebra tools.\footnote{Because physical phenomena
  are often continuous and digital computation is often discrete, it
  is important that such languages support hybrid
  (continuous/discrete) systems.  Part I presents examples of hybrid
  models, including examples of modeling quantization and
  discretization. The focus of this paper is on issues that relate
  only to continuous models.}

\subsection{The Tool Chain Coherence Challenge}
Based on our experience in several domains, it appears that scientists
and engineers engaged in CPS design are often forced to transfer
models through a chain of disparate, specialized design tools. Each
tool has a clear purpose. For example, drafting tools like AutoCAD and
Solidworks support creating images of new products; MATLAB, R, and
Biopython support the simple programming of dynamics and control; and
finally, tools like Simulink support the simulation of device
operation ~\cite{zhu2010mathematical}. Between successive tools in
this chain, there is often little or no formal communication. It is
the responsibility of the CPS designer to translate any data or
valuable knowledge to the next tool.

We believe that tool chain coherence should be approached from a
linguistic point of view.  Precise reasoning about models during each
step of the design process can readily be supported by applying
classical (programming) language design principles, including defining
a formal semantics.  Reasoning across design steps can be facilitated
by applying two specific ideas from language design: 1) increasing
the expressivity of a language to encompass multiple steps in the
design process; and 2) automatically compiling models from one step
to the next.  The latter idea reduces manual work and eliminates
opportunities for mistakes in the translation.

\subsection{Recapitulation of Part I}\label{sec-vis}

Part I of this work~\cite{taha2013modeling} identifies a set of
prominent aspects that are common to CPS design, and shows the extent
to which a small core language, which we call Acumen, supports the
expression of these aspects. The earlier paper considers the
following aspects:
\begin{enumerate}
\item Geometry and visual form
\item Mechanics and dynamics
\item Object composition
\item Control
\item Disturbances
\item Rigid body dynamics
\end{enumerate}
and presents a series of examples illustrating the expressivity
and convenience of the language for each of these aspects.  In this narrative,
the last aspect stands out as more open-ended and potentially
challenging.  The early paper considers a single case study involving a
rigid body: a single link rod (two masses connected by a fixed-length
bar).  Thus, it does not address the issue of how well such a small
language is suited to modeling larger rigid body systems.

\subsection{Contributions}

Modeling continuous dynamics is only one aspect of CPS designs that we
may need to model, but it is important because it is particularly
difficult.  This paper extends previous work by considering the
linguistic demands posed by two larger case studies drawn from the
rigid body domain.  After a brief review of Acumen
(Section~\ref{sec-Acumen}), the first case study we consider is a
quadcopter, which is a complex, rigid body system that is often used
as a CPS example.  The quadcopter case study shows that Acumen can
simply and directly express Newtonian
models (Section~\ref{sec-Quadcopter}).  The second case study is a
research robot called the RiceWrist-S.  In this case, developing a
Newtonian model is difficult and inconvenient.  It illustrates how the
more advanced technique of Lagrangian modeling can be advantageous for
some problems. As a prelude to modeling the Rice Wrist-S robot, we
consider two ostensibly simple dynamic systems, namely a single
pendulum and a double pendulum.  For the second system, we show that
Lagrangian modeling leads to a much simpler mathematical
formalization (Section~\ref{sec-Analytical}).  To confirm this
insight, we show how Lagrangian modeling enables us to construct a
simple model of the dynamics for the RiceWrist-S
(Section~\ref{sec-RWS}).  This analysis provides stronger evidence of
the need to support partial derivatives and implicit equations in any
hybrid-systems language that is expected to support the rigid body
systems domain (Section~\ref{sec-Discussion}).\footnote{The original
  design for Acumen supported partial derivatives
  \cite{zhu2010mathematical}.  So far, partial derivatives have been
  absent from the new design \cite{taha-2012}.  This paper develops
  the rational for this construct more systematically, with the goal
  of justifying its introduction in a future revision of Acumen.}

\/*To address these questions, this paper identifies a set of
prominent CPS aspects that are common, and shows the extent to which
the core language described above can support expressing these
aspects.  Specifically, we consider the following aspects:

\begin{itemize}
\item Geometry and visual form (Section \ref{sec-vis})
\item Particle dynamics (Section \ref{sec-mech})
\item Composite objects (Section \ref{sec-obj}) 
\item Control (Section \ref{sec-control})
\item Rigid body dynamics (Section \ref{sec-rigid})
\end{itemize}
We also show how the core language described above expresses
archetypical examples of each of these aspects. */

\section{Acumen}\label{sec-Acumen}

Modeling and simulation languages are an important class of
domain-specific languages.  For a variety of reasons, determining what
are the desirable or even plausible features in a language intended
for modeling and simulation of hybrid systems~\cite{carloni-2006} is
challenging.  For example, there is not only one notion of hybrid
systems but numerous: hybrid systems, interval hybrid systems,
impulsive differential equations (ordinary, partial), switching
systems, and others.  Yet we are not aware of even one standard
example of such systems that has a simple, executable semantics.  To
overcome this practical difficulty in our analysis, we use a small
language called Acumen~\cite{taha-2012,acumen-web}, and assign it a
simple, contant time step, semantics.  We are developing this
language to apply the linguistic approach to the accessibility and
tool chain coherence challenges identified above.

The language consists of a small number of core constructs, namely:
\begin{itemize}
\item Ground values (e.g., \verb|True|, \verb|5|, \verb|1.3|, \verb|"Hello"|)
\item Vectors and matrices (e.g., \verb|[1,2]|, \verb|[[1,2],[3,4]]|)
\item Expressions and operators on ground and composite types 
      (\verb|+|, \verb|-|, ...)
\item
  Object class definitions (\verb|class C (x,y,z)| \verb| ... end|)
\item
  Object instantiation and termination operations (\verb|create|, \verb|terminate|)
\item
  Variable declarations (\verb|private ... end|).  For convenience, we
  included in the set of variables a special variable called
  \verb|_3D| for generating 3D animations.
\item
  Variable derivatives (\verb|x'|, \verb|x''|, ...) with respect to time
\item
  Continuous assignments (\verb|=|)
\item
  Discrete assignments (\verb|:=|)
\item
  Conditional statements
  (\verb|if|, and 
   \verb|switch|)
\end{itemize}
It should be noted that derivatives in this core language are only
with respect to an implicit variable representing global time.  In
this paper we will consider concrete examples illustrating why it will
be useful (in the future) to introduce partial derivatives to Acumen.

Continuous assignment is used to express differential equations,
whereas discrete assignments are used to express a discontinuous
(sudden) change.  Initial values for variables (at the time of the
creation of a new object) are specified using discrete assignments.
Currently, we take a conservative approach to initial conditions,
which require users to express them explicitly even for variables
where there is a continuous equation that will immediately override
this explicit initial value.  Finaly, for the sake of minimality,
Acumen has no special notation for introducing constants (in the sense
of variables that do not change value over time).

We are using this language for a term-long project in a course on CPS
\cite{taha-lecture-notes}, which has been enthusiastically received in
the first two offerings of this course (see for example
\cite{TahaWESE13, TahaCPScourse}). A parsimonious core language can
help students see the connections between different concepts and avoid
the introduction of artificial distinctions between manifestations of
the same concept in different contexts.  This bodes well for the
utility of such languages for addressing this challenge.  However, to
fully overcome this challenge, we must develop a clear understanding
of how different features in such a language match up with the demands
of different types of cyber-physical systems.

The Acumen distribution contains implementations of multiple different
solvers for simulation, accessible from a ``Semantics'' menu.  For
this paper, we use on the ``Traditional'' semantics, which simply uses
Eulers method and a constant time step for integration.

{\bf Remark about Syntax:} Since the writing of Part I of this paper,
a minor change has been made to the syntax, where \verb|:=| now
describes discrete assignments, and \verb|=| now describes continuous
assignments. Also, a syntax highlight feature has been introduced, in
order to improve the user experience.

\/*
\section{Geometry and Visual Form}

Essentially all physical systems either take up space or have an
effect on space.  As a result, visual presentation plays an essential
role in the CPS design. For many people, it is hard to imagine a
design without conjuring an image of a general visual form. If we want
to replace physical prototyping with virtual prototyping,
visualization becomes a necessity. Animating the evolving state of a
hybrid system using visual geometric presentation often reveals
behaviors of the system that might otherwise be undetected. From a
pedagogic point of view, the trigonometric reasoning involved in
creating visualizations motivates studying the geometry of motion
(kinematics).

\subsection{Drawing 3D Objects}
A hybrid modeling and simulation language can be naturally extended
with a lightweight mechanism for three dimensional (3D) visualization
\cite{Yingfu-Thesis}.

In the core language, the user can specify 3D visualizations through a
special variable called \verb|_3D|.  This variable is special only in
that it is read by the implementation and used to generate a dynamic
3D scene.  In principle, any graphical rendering technology can be
used by an implementation to realize these visualizations.  In
practice, the current implementation uses the Java3D library, which is
built on top of OpenGL~\cite{shreiner2009opengl}.

\begin{figure}
  \centering
  \includegraphics[width=0.7\columnwidth]{walid/exampleG1}
  \caption{
    The 3D output generated for an instance of the class sphere.
  }\label{fig:sphere-scr}
\end{figure}

\subsection{Class Definitions and Parameterization}

The following class definition specifies a particular way for drawing
a sphere:
\begin{verbatim}
class sphere (m,D)
 private p :=[0,0,1];_3D := [] end
 _3D = [["Sphere", D+p, 0.03*sqrt(m),
           [m/3,2+sin(m),2-m/2], [1,1,1]]]
end
\end{verbatim}
The private variable \verb|m| represents a mass.  This parameter is
only used to pick a size and a color for the sphere.  The parameter
\verb|D| is a display reference point.  Passing different \verb|D|
values to individual objects facilitates creating visualizations where
the individual objects appear in different places.  The \verb|private|
section declares variables present in each object as well as their
initial values at the (simulated) time when an object is created.  The
variable \verb|p| is used to represent the position of the sphere. The
special variable named \verb|_3D| must be bound to a vector with a
format determined by the 3D visualization system used in of our
implementation of the core language.  The continuous assignment is
computed for as long as the object exists in simulation. The
\verb|_3D| vector has the following format. The first field is a
string indicating that the shape we want is a sphere.  The second
field is the coordinate for the center of the sphere.  The third field
is the radius.  In Figure~{\ref{code}}, we chose to make the radius a
simple function of the mass.  This function is not intended to have
any physical meaning other than to produce reasonable effects for the
examples presented in this paper.  The next field is a vector that
represents the red/green/blue (RGB) colors for this sphere. In
Figure~{\ref{code}}, we used an {\em ad hoc} formula to generate a
color based on the mass.  The last field of this vector specifies the
orientation of the object, which only matters when the sphere has a
texture. \Fig{sphere-scr} depicts a visualization generated using this
class.

\subsection{Object Creation, Continuous Assignment, and Animation} We can create
a sphere by writing \verb|s := create sphere| \verb|(5,[0,0,0])| in the
initialization (private) section and then \verb|s.p =| \verb|[0.1, 0.2, 0.3]| in
the body.  To generate 3D animations, all we have to do is to let the value of
\verb|p| vary over time, as in the following code: \begin{verbatim} class
moving_sphere (m,D) private s := create sphere (m,D); t := 0; t' := 0; end t'  =
5; s.p = [sin(t)*sqrt(1-(sin(t/10)^2)), cos(t)*sqrt(1-(sin(t/10)^2)),
sin(t/10)]; end \end{verbatim} Here the variable \verb|t| and its derivative
\verb|t'| are introduced to model a local variable that progresses at exactly
five times the rate of time.  All that is needed to accomplish this is to
include the equation \verb|t'=5|.  The time-varying variable \verb|t| is then
used to generate some interesting values for the X, Y, and Z components of the
position field \verb|p| that represents the center of the sphere object
\verb|s|.

As noted earlier, we can create multiple instances of a visualization class
(such as \verb|moving_sphere|) at different locations by varying the \verb|D|
parameter.  By changing the value of the position parameter \verb|p|, we can
create an animation with two spheres moving in a synchronized fashion.

It is useful to note that a 3D visualization facility can be used
to visualize not only 3D values but also scalar values.  For example,
it is useful to define objects that assist in visualizing specific
scalar values during a simulation.  The following code defines a
class to visualize a scalar value as a cylinder, whose length is proportional
to that value:

\begin{verbatim}
class display_bar (v,c,D)
 private _3D := [] end
 _3D = ["Cylinder", D+[0,0.2,v/2], [0.02,v],c,
        [-pi/2,0,0]];
end
\end{verbatim}
Following the string \verb|Cylinder|, the next argument represents the
center of the cylinder.  We take this to be \verb|v/2| because this will
allow us to keep one end of the cylinder fixed as the value of \verb|v|
changes.  The next argument is a tuple containing the radius and
length of the cylinder.  The next argument is color.  The last
argument specifies orientation angles for the cylinder.

\subsection{Vector and Trigonometric Calculation} In many cases, it is necessary
to perform a bit of geometrical calculation to create the desired shape. Such
calculations may be required in simple situations. An example of such a context
is drawing a cylinder between two points.  Often, visualizations cannot be
specified directly because the underlying library uses a different approach to
describe the orientation of a figure.  In the case of cylinders, it is common to
use polar coordinates (two angles) to specify the orientation of the axis of a
cylinder. Once we have figured out all necessary calculations, they can be
encapsulated in one class as follows:

\begin{verbatim}
class cylinder (D)
 private 
  p :=[0,0,0]; q:=[0,0,0]; radius := 0.01;
  length := 0.01; alpha:=0; theta:= pi/2;
  x:=0;y:=0;z:=0; _3D := [] 
end
 x = dot(p-q,[1,0,0]); y = dot(p-q,[0,1,0]);
 z = dot(p-q,[0,0,1]);
 length = norm(p-q); alpha = asin(z/length);
 if (y>0)
   theta = asin(x/(length*cos(alpha)))
 else
   theta = -asin(x/(length*cos(alpha)))+pi
 end
 _3D = [["Cylinder",(p+q)/2+D,[radius,length],
           [1,1,1],[alpha,0,-theta]]];
end
\end{verbatim}
The \verb|dot| and \verb|norm| operators compute the dot product and
the vector norm (or length).  Creating such an object is a good
exercise in making customized building blocks for visualization.

\section{Particle Dynamics}\label{sec-mech}

The most basic approach to model the mechanical dynamics of a system
is to view it as a point mass, or a {\em particle}.  In contrast to
the syntax needed to describe geometric and visual objects, describing
particle dynamics can be done more
concisely.  

\subsection{Transitioning from 1D to 3D with Ease}\label{sec-trans} 
A point mass that can only move in one
dimension can be represented as follows: \begin{verbatim} class
  mass_1d (m,p0,D) private p:=p0; p':=0; p'':=0; f:=0; e_k:=0;
  s:=create sphere (m,D) end p'' = f/m; e_k = 0.5 * m * (p') ^2; s.p =
  [0,0,p] end \end{verbatim} The object constructor takes three *
parameters: a mass \verb|m,| an initial position \verb|p0|, and a
reference * point for visualization.  Internally, the mass keeps track
of a position * \verb|p|, its first and second derivatives \verb|p'|
and \verb|p''|, a force * \verb|f|, and the kinetic energy \verb|e_k|.
For visualization, a sphere * object is created during initialization.
The body of the class definition * specifies that the acceleration of
the object, \verb|p''|, is determined by * Newton's law $F=ma$, where
we are solving for acceleration (which is just * \verb|p''| here).
The expression for energy uses the built-in dot-product * operation on
vectors.  Finally, we set the position \verb|p| of the visual * object
sphere to be the same as the position \verb|p| of the current object.

Supporting vector operations makes it possible to define a similar object that
has a three dimensional position almost as simply: \begin{verbatim} class mass
(m,p0,D) private p:=p0; p':=[0,0,0]; p'':=[0,0,0]; f:=[0,0,0]; e_k:=0; s :=
create sphere (m,D); end p'' = f/m; e_k = 0.5 * m * (dot(p',p')); s.p = p; end
\end{verbatim} Note that it is convenient in this domain to compute derivatives
of vectors.  We can induce continuous behaviors in such an object by means of an
external continuous assignment.  For example, the effect of a gravitational
force on a mass object \verb|m| by a continuous assignment \verb|m.f =
m.m*[0,0,-9.81]|.  An idealized, 3D spring can be modeled as follows:
\begin{verbatim} class spring (k,l0,D) private p1:=[0,0,0]; p2:=[0,0,0];
f1:=[0,0,0]; f2:=[0,0,0]; dl := [0,0,0]; e_p:=0; end dl  = p2-p1 *
(1-l0/norm(p2-p1)); f1  =  k*dl; f2  = -k*dl; e_p = 0.5 * k * dot(dl,dl); end
\end{verbatim} This class associates a different force with each end of the
spring. It computes a potential energy \verb|e_p| rather than a kinetic energy.
No visualization is included in this object, but that can be easily done using
the techniques presented above.

\subsection{Impacts and Discrete Assignment}
An important physical effect in dynamics is impact, often modeled as a
 sudden effect. Therefore, discrete assignments can be used for this
 purpose.  The following model provides an example of the use of
 discrete assignment to model the impact of a falling ball against a
 floor:
\begin{verbatim}
class bouncing_ball (D)
 private
  m  := create mass_1d (10, 3, D);
  bk := create display_bar 
        (0,[3, 0.2, 0.2], D+[ 0.1, 0.2, 0]);
  bp := create display_bar 
        (0,[0.2, 3, 0.2], D+[-0.1, 0.2, 0]);
  bt := create display_bar 
        (0,[0.2, 0.2, 3], D+[   0, 0.2, 0]);
 end
 m.f = m.m * -9.81;
 if (m.p < 0 && m.p' < 0)
  m.p' := -0.9 * m.p'
 end;
 bk.v = m.e_k / (m.m * 9.81);
 bp.v =   (m.m * 9.81 * m.p) / (m.m * 9.81);
 bt.v = bk.v + bp.v;
end
\end{verbatim}
The model uses the mass class along with a continuous gravity model
and a ground-impact model where the ball loses 10\% of its velocity.
The class \verb|display_bar| is used to display colored bars to present some
additional information in the 3D output.  The mass model used here has
only one degree of freedom along the Z axis.  We use three display
bars to visually represent the kinetic and potential energy, as well
as their sum.  The discrete assignment occurs inside the if statement
that detects impact with the ground plane.  \Fig{bball-scr} shows a
sequence of screenshots, one including the Integrated Development
Environment (IDE), which results from running this example.  It can be
seen that, as expected, the total energy decreases at each impact,
while the kinetic and potential energies reach their respective maxima
and minima at the height of the bounce and the impact at ground level.
\begin{figure}
  \centering
  \includegraphics[width=\columnwidth]{rolo/bball-ide-scr1-cropped}\\[1.1\baselineskip]
  \includegraphics[width=0.3\columnwidth]{rolo/bball-ide-scr2-cropped}\hfill
  \includegraphics[width=0.3\columnwidth]{rolo/bball-ide-scr3-cropped}\hfill
  \includegraphics[width=0.3\columnwidth]{rolo/bball-ide-scr4-cropped}
  \caption{The IDE of the implementation with the bouncing ball model
    and simulation results.  The green bar indicates the potential
    energy, the red one is the kinetic energy, and the blue bar is
    their sum.  The total energy decreases with each ground impact,
    and during the free flight phase the two energies behave as
    expected.  }\label{fig:bball-scr}
\end{figure}
Now we turn to creating systems made from components, such as the mass
and spring examples, that we have just introduced.

\section{Composite Objects}\label{sec-obj} Most systems are composite, in that
they consist of smaller, interacting components.  The need to express such
compositions is the motivation for supporting a class system in the proposed
core language.  Connecting components is a matter of relating fields in
different components though continuous assignments. 

For example, the following class models a system consisting of three
masses connected by two springs. Note that the class uses an instance
of the class \verb|display_bar| to draw a cylinder to display the
kinetic energy in the system.
\begin{verbatim}
class example_3 (D)
 private 
  m1 := create mass (15,[0,0, 1],D);
  m2 := create mass (5, [0,0,-1],D);
  m3 := create mass (1, [0,0,-1.5],D);
  s1 := create spring (5,1.75,D);
  s2 := create spring (5,0.5,D);
  b  := create display_bar (-1.5,0,D)
 end
 s1.p1 = m1.p; s1.p2 = m2.p; s2.p1 = m2.p;
 s2.p2 = m3.p; m1.f  = s1.f1;
 m2.f  = s1.f2 + s2.f1; m3.f  = s2.f2;
 b.v   = (m1.e_k + m2.e_k + m3.e_k
            + s1.e_p + s2.e_p)*12;
end
\end{verbatim}
\section{Control}\label{sec-control}

Almost every CPS has a control aspect.  The goal of control is to apply a
manipulated variable to a system to bring a measured variable close to a desired
goal. Fundamentally, many systems that we care to model can be described in
terms of rigid body models, so it is essential that our core language can
effectively handle them. In the context of the model presented above, and given
a controller object \verb|c|, the introduction of such a controller can be
modeled as follows:

\begin{verbatim}
// Goal is spring length at rest
 c.g = s1.l+s2.l;
// Value is actual spring length
 c.v = m1.p-m3.p;
 // Add c.f
 m1.f = s1.f1 + c.f;  m2.f = s1.f2 + s2.f1;
// Subtract c.f
 m3.f = s2.f2 - c.f;
\end{verbatim}
In this model the goal value for the controller is to have the length
of the system be the same as the uncompressed lengths of the two springs.
The quantity that we wish to control is the position of the first mass
minus the position of the third one.  The way we will achieve that is
to take a force value \verb|f| that is generated by the controller and apply
it to both sides of the system that we have constructed, but in
opposing directions.
\subsection{PID Control}
Now the question that remains is how the controller \verb|c| should compute
its output force \verb|f| given the goal \verb|g| and measured value \verb|v|. This is a
prototypical question in the design of control systems, and one that can
be approached in a variety of different ways. The most basic type of controller
is a proportional-integral-derivative (PID) feedback controller. Often, all
three forms of feedback are not required to ensure the level of stability or
performance required, and it is up to the designer to determine the best mix.
The first term of the force \verb|f| computed is the proportional term, and is
directly proportional to the difference between the goal \verb|g| and current
value \verb|v| of the quantity that we want to control.  The second term of the
control force, the integral term, exerts a higher force only after a weaker
force has been tested for some time. This can be helpful, for example, if there
are external constant forces (such as gravity) acting on our system, and we do
not know their precise quantity ahead of time. The third term of the control
force, the derivative term, is modulated based on the rate of change of the
value being measured and can therefore deal with oscillatory behavior.  A PID
controller with these behaviors is modeled below.  \begin{verbatim} class
force_controller_pid (k_p,k_i,k_d) private g:=[0,0,0]; v:=[0,0,0]; s:=[0,0,0];
f:=[0,0,0]; i:=[0,0,0]; i':=[0,0,0] end f = k_p*(g-v) + k_i*i - k_d*s; i' =
(g-v) end \end{verbatim} Note that this controller has an extra field \verb|s|
that should be provided from outside the object to serve as the derivative of
the error term that should affect the final force \verb|f|. The variable
\verb|i| is being used to integrate the difference between the goal \verb|g| and
the value \verb|v| over time, so no extra inputs are needed.

To test our controller design, it can be useful to introduce various
sources of disturbance into the system.  At least for preliminary
experimentation, it can be sufficient to model such disturbances as
autonomous sources of various forces. These sources can be added to
the system in much the same, simple way that the controllers were
added previously.

\subsection{Discretization and Quantization}\label{sec-discretization} One
aspect of controllers not captured in the preceding models is the fact that they
are generally implemented as code executing on a digital computer.  As a result,
the actual behavior of controllers is not identical to the simple mathematical
model presented in the preceding section.  In fact, the computed control force f
is affected both by the discretization of time and quantization in the
representation of physical quantities. Both effects can be concisely expressed
in the core language.  To model discretization, the key mechanism needed is to
define a local clock and to only allow actions to be performed (or to be
observed) at clock transitions.  The following class models a PID controller
(like the one presented above) with discretization and quantization effects.
\begin{verbatim} class force_controller_pid_d (k_p,k_i,k_d,period) private
g:=[0,0,0]; v:=[0,0,0]; s:=[0,0,0]; f:=[0,0,0]; t:=0; t':=0; i:=[0,0,0];
i':=[0,0,0] end t' = 1; if (t>period) t:=0; f  = k_p*(g-v) + k_i*i - k_d*s; end;
i' = (g-v) end \end{verbatim} The variable \verb|t| and its derivative \verb|t'|
are used in a similar manner to that used at the start of this paper to generate
an interesting signal for \verb|moving_sphere|.  Here we do two new things with
the variable \verb|t|.  The first is that we have a conditional statement based
on this variable that waits until \verb|(t>period)|.  The parameter
\verb|period| models the time it takes the particular microprocessor that
implements our controller to produce the new value of the result of the
controller.  Once the condition is true, the next thing we do is to reset the
counter.  Then, we reset its value to \verb|0| using the statement \verb|t:=0|
as soon as that condition is true.  In addition to this reset, the conditional
also allows the equation for variable \verb|f| in the original model to take
effect only for that instant when \verb|t| has surpassed the value of
\verb|period|.  Because no other definition is given for this value until this
event occurs again (at the start of the next period), the value \verb|f| remains
constant until that change occurs. With this model, it is easy to illustrate
that, as the sampling period goes up, the system that we are trying to control
can become unstable.
*/

\section{Quadcopter}\label{sec-Quadcopter} 

A rigid body system consists of a set of solid bodies with
well-defined mass and inertia, connected by constraints on distances
and/or angles between the solid bodies.  The dynamics of many physical
systems can be modeled with reasonable accuracy as a rigid body
system.  It is widely used for describing road vehicles, gear systems,
robots, etc.  In this section, we consider an example of a complex
system that can be successfully modeled as simple rigid body, namely,
the quadcopter.

\subsection{Background}
The quadcopter is a popular mechatronic system with four rotor blades
to provide thrust. This robust design has seen use in many UAV
applications, such as surveillance, inspection, and search and
rescue. Modeling a quadcopter is technically challenging, because it
consists of a close combination of different types of physics,
including aerodynamics and mechanics.  A mathematical model of a
quadcopter may need to address a wide range of effects, including
gravity, ground effects, aerodynamics, inertial counter torques,
friction, and gyroscopic effects.

\subsection{Reducing Model Complexity Through Control}
Even if we limit ourselves to considering just six degrees of freedom
(three for position and three for orientation), the system is
underactuated (one actuation from each rotor vs. six degrees of
freedom) and is therefore not trivial to control.  Fortunately, controllers exist that can ensure that actuation is realized by getting 
the four rotors to work in pairs, to balance the forces and torques of the system.  With this approach, the quadcopter
can be usefully modeled as a single rigid body with mass and inertia,
by taking account of abstract force, gravity and actuation control
torques.  This model is depicted in Fig.~\ref{fig:quad-fbd}.

\/* Quadcopter dynamics are
well understood and can be modeled in several ways ({\em cf}
~\cite{quadA11,quadCarrillo}).*/

\subsection{Mathematical Model}
\begin{figure}
  \centering
  \includegraphics[width=0.8\columnwidth]{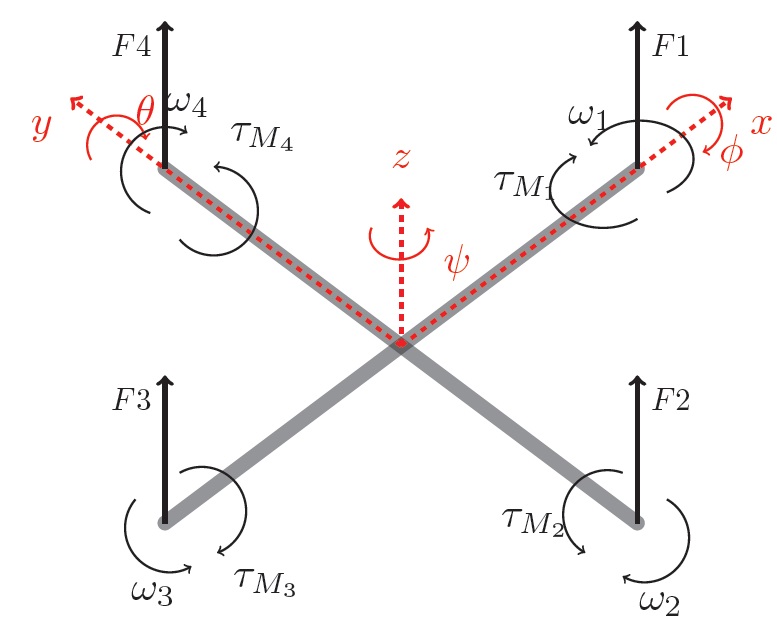}
  \caption{Free body diagram of the quadcopter}
  \label{fig:quad-fbd}
\end{figure}

To generate the equations for the dynamics of our common quadcopter
model~\cite{quadA11}, we first construct the rotational matrix to
translate from an inertial (globally-fixed) reference frame to the
body-fixed reference frame shown in Fig.~\ref{fig:quad-fbd}. This
matrix represents rotation about the $y$ axis ($\theta$), followed by
rotation about the $x$ axis ($\phi$), and then rotation about the $z$
axis ($\psi$).
\begin{equation}
R=
\begin{bmatrix}
C_\psi C_\theta & C_\psi S_\theta C_\phi -S_\psi C_\phi & C_\psi S_\theta C_\phi+S_\psi S_\phi\\
S_\psi C_\theta & S_\psi S_\theta S_\phi + C_\psi C_\theta & S_\psi S_\theta C_\phi - C_\psi S_\phi \\
-S_\theta & C_\theta S_\phi &C_\theta C_\phi
\end{bmatrix}
\end{equation}
Here, $C$ , $S$ and $T$ refer to $cos$, $sin$, and $tan$, respectively. Next, summing forces on the quadcopter results in:
\begin{equation}
\sum F = m\bar{a} = G + RT
\end{equation}
where $G$ is the force due to gravity, $R$ is the rotational matrix, and $T$ is the thrust from the motors. This expands to 
\begin{equation}
\begin{bmatrix}
\ddot{x}\\
\ddot{y}\\
\ddot{z}
\end{bmatrix}
=-g 
\begin{bmatrix}
0\\
0\\
1
\end{bmatrix}
+
\frac{T}{m}
\begin{bmatrix}
C_\psi S_\theta C_\phi+S_\psi S_\phi\\
S_\psi S_\theta C_\phi - C_\psi S_\phi \\
C_\theta C_\phi
\end{bmatrix}
\end{equation}
Finally, by summing moments about the center of mass, the equations
for the dynamics for each of the rotational degrees of freedom can be
determined as follows:

\begin{multline}
\begin{bmatrix}
\ddot{\phi}\\
\ddot{\theta}\\
\ddot{\psi}
\end{bmatrix}
=
\begin{bmatrix}
0 & \dot{\phi}C_\phi T_\theta +\dot{\theta}\frac{S_\phi}{C_\theta ^2} &
-\dot{\phi}S_\phi C_\theta +\dot{\theta}\frac{C_\phi}{C_\theta ^2}\\
0 & -\dot{\phi}S_\phi & -\dot{\phi}C_\phi\\
0 & \dot{\phi}\frac{C_\phi}{C_\theta}+\dot{\phi}S_\phi
\frac{T_\theta}{C_\theta} &
-\dot{\phi}\frac{S_\phi}{C_\theta}+\dot{\theta}C_\phi \frac{T_\theta}{C_\theta}
\end{bmatrix}\nu
\\
\\ +W^{-1}_\eta \dot{\nu}
\end{multline}
Where

\begin{equation}
\nu=
\begin{bmatrix}
p\\q\\r
\end{bmatrix}
= W_\eta 
\begin{bmatrix}
\dot{\phi}\\
\dot{\theta}\\
\dot{\psi}
\end{bmatrix}
=
\begin{bmatrix}
1 & 0 & -S_\theta \\
0 & C_\phi & C_\theta S_phi\\
0 & -S_\phi & C_\theta C_\phi
\end{bmatrix}
\begin{bmatrix}
\dot{\phi}\\
\dot{\theta}\\
\dot{\psi}
\end{bmatrix}
\end{equation}
\begin{equation}
\dot{\nu}=
\begin{bmatrix}
\left(I_yy-I_zz\right)\frac{q r}{I_xx}\\
\left(I_zz-I_xx\right)\frac{q r}{I_yy}\\
\left(I_xx-I_yy\right)\frac{q r}{I_zz}\\
-I_r
\end{bmatrix}
\begin{bmatrix}
\frac{q}{I_xx}\\
\frac{-p}{I_yy}\\
0
\end{bmatrix}
\omega_\Gamma+
\begin{bmatrix}
\frac{\tau_\phi}{I_xx}\\
\frac{\tau_\theta}{I_yy}\\
\frac{\tau_\psi}{I_zz}
\end{bmatrix}
\end{equation}
\begin{equation}
\begin{bmatrix}
\tau_\phi\\
\tau_\theta\\
\tau_\psi
\end{bmatrix}
=
\begin{bmatrix}
l k \left(-\omega^2_2 +\omega^2_4\right)\\
l k \left(-\omega^2_1 +\omega^2_3\right)\\
\sum_{i=1}^{4} \tau_{M_{i}}
\end{bmatrix}
\end{equation}

\subsection{Acumen Model for Quadcopter}\label{app-quad}
\begin{figure} \centering
\includegraphics[width=0.65\columnwidth]{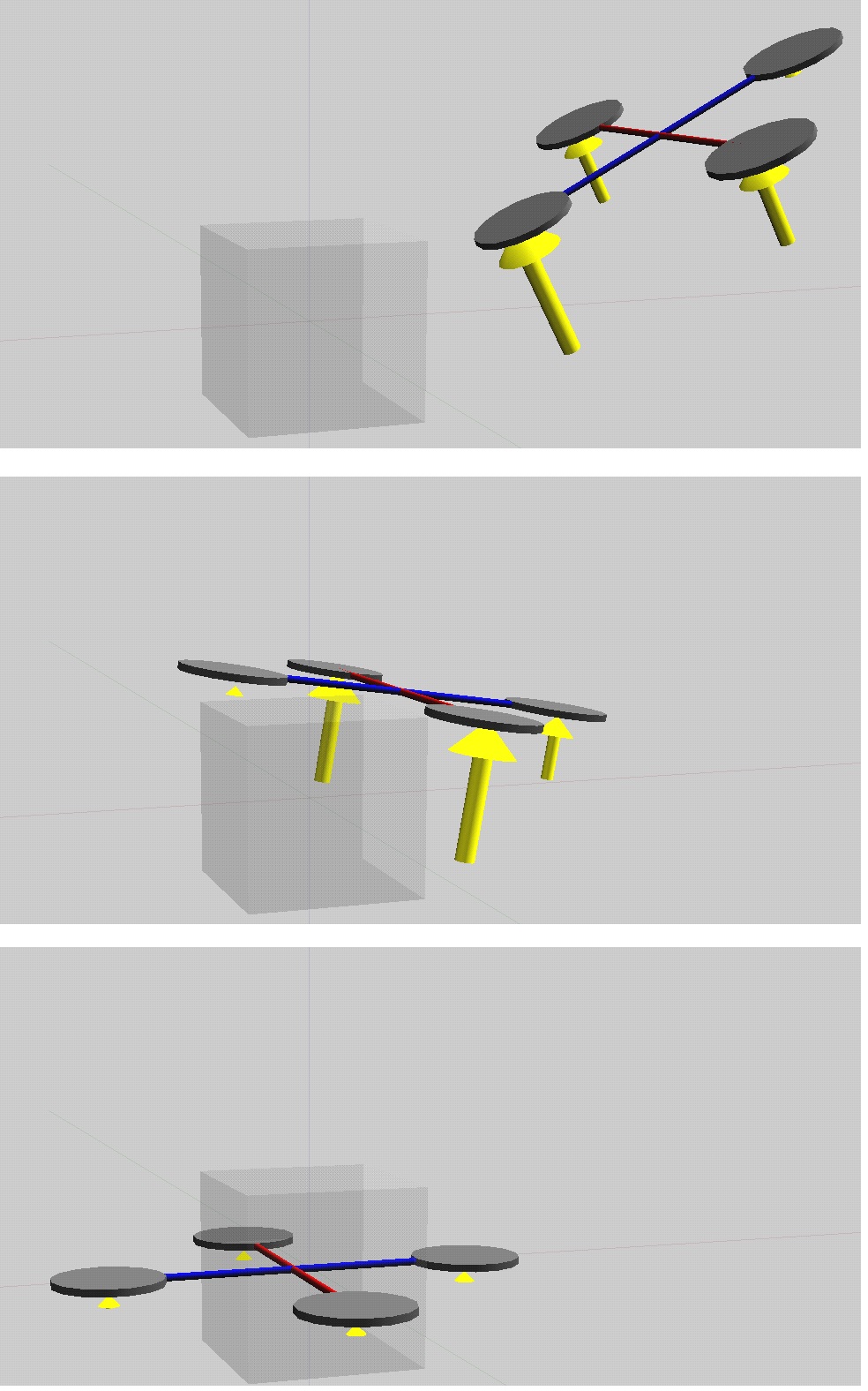} \caption{The
  simulation results of the quadcopter model with PID control.  Here
  the controller is bringing the quadcopter elevation, roll, pitch,
  and yaw to zero.  Yellow arrows attached to each rotor indicate the
  thrust force generated by the
  rotors.}\label{fig:quad-scr} \end{figure}

The equations derived earlier for the dynamics can be expressed in our
core language as follows:
\begin{verbatim}
class QuadCopter(P,phi,theta,psi)
private
  g := 9.81;     m := 0.468;                
  l := 0.225;    k := 2.98*10^(-6);
  b := 1.140*10^(-7); 
  IM := 3.357*10^(-5);  
  Ixx := 4.856*10^(-3); 
  Iyy := 4.856*10^(-3);    
  Izz := 8.801*10^(-3);  
  Ax  := 0.25;  Ay  := 0.25; 
  Az  := 0.25; 
  w1 := 0;  w2:= 0;   w3 := 0;                  
  w4 := 0;  wT := 0;  f1 := 0; 
  f2 := 0;  f3 := 0;  f4 := 0;                  
  TM1 := 0; TM2 := 0; T := 0; 
  TM3 := 0; TM4 := 0;
  P' := [0,0,0];    P'' := [0,0,0];
  phi' := 0; theta' := 0; psi' := 0;
  phi'' := 0; theta'' := 0; psi'' := 0; 
  p := 0; q := 0; r := 0; p' := 0;      
  q' := 0; r' := 0; Ch:=0; Sh:=0;
  Sp:=0; Cp:=0; St:=0; Ct:=0; Tt:=0
end
  T = k* (w1^2 + w2^2 + w3^2 + w4^2);
  f1 = k * w1^2; TM1 = b * w1^2;
  f2 = k * w2^2; TM2 = b * w2^2;
  f3 = k * w3^2; TM3 = b * w3^2;
  f4 = k * w4^2; TM4 = b * w4^2;
  wT = w1 - w2 + w3 - w4;

  Ch = cos(phi); Sh = sin(phi);
  Sp = sin(psi); Cp = cos(psi);
  St = sin(theta); Ct = cos(theta); 
  Tt = tan(theta);

  P'' = -g * [0,0,1] + T/m
   *[Cp*St*Ch+Sp*Sh,Sp*St*Ch-Cp*Sh,Ct*Ch] 
   -1/m*[Ax*dot(P',[1,0,0]),
         Ay*dot(P',[0,1,0]),
         Az*dot(P',[0,0,1])];          
  p' = (Iyy-Izz)*q*r/Ixx - IM*q/Ixx*wT
   + l*k*(w4^2 - w2^2)/Ixx;
  q' = (Izz-Ixx)*p*r/Iyy-IM*(-p)/Iyy*wT
   + l*k*(w3^2 -w1^2)/Iyy;
  r' = (Ixx - Iyy)*p*q/Izz  + b*(w1^2  
   + w2^2 -w3^2 -w4^2)/Izz;
  phi'' = (phi'*Ch*Tt+ theta'*Sh/Ct^2)*q 
   + (-phi'*Sh*Ct+theta'*Ch/Ct^2)*r 
   + (p'+q'*Sh*Tt+r'*Ch*Tt);
  theta'' = (-phi'*Sh)*q + (-phi'*Ch)*r 
   + (q'*Ch+r'*(-Sh));
  psi'' = (phi'*Ch/Ct+phi'*Sh*Tt/Ct)*q 
   + (-phi'*Sh/Ct+theta'*Ch*Tt/Ct)*r 
   + (q'*Sh/Ct+r'*Ch/Ct);
end
\end{verbatim}
 ~\Fig{quad-scr} presents snapshots
of a 3D visualization of the quadcopter responding to a
signal from a basic stabilizing controller \cite{quadA11}. This
example shows that the Acumen core language can express models for complex
mechatronic systems that are widely used in both research and education today.

\section{Lagrangian Modeling, and Why we Need it}\label{sec-Analytical}

Mathematical modeling of rigid body systems draws heavily on the field
of classical mechanics.  This field started in the 17th century with
the introductions of Newton's principles of motion and Newtonian
modeling.  Newton's foundational work was followed by Lagrangian
modeling in the 18th century, and Hamiltonian modeling in the 19th.
Today, mechanical engineers make extensive use of the Lagrangian
method when analyzing rigid body systems.  It is therefore worthwhile
to understand the process that engineers follow when using this
method, and to consider the extent to which a modeling language can
support this process.

The Newton method is focused on taking into consideration the forces
and torques operating on a rigid body, and then computing the linear
and angular accelerations of the center of mass of that rigid body. In
general, this method consists of isolating the rigid body of interest
in a free body diagram, selecting coordinate frame and summing forces
and torques on the rigid body with respect to that frame, then using
kinematics to express the linear and angular acceleration terms,
before finally deriving the equations for the dynamics.

\hide{
From the accessibility point of view, it is noteworthy that it
illustrates some of the benefits of using vector algebra to model and
reason about the dynamics of systems.  In particular, vector algebra
can often allow us to think about problems in 2D and then have the
results generalize naturally to 3D, much like the 1D to 3D
transformation (Section \ref{sec-trans}).}

The Lagrangian method is based on the notion of {\em action} $L=T-V$,
which is the difference between the kinetic $T$ and potential energy
$V$.  The Euler-Lagrange equation is itself a condition for ensuring
that the total action in the system is stationary (constant).  In
Lagrangian modeling of physical systems, this condition should be seen
as the analogy of the combined conditions $\Sigma F = ma$ and $\Sigma
\tau = I\omega''$ in Newtonian mechanics.  The Euler-Lagrange equation
is as follows:
\begin{equation}\label{eq-el}
\forall i\in\{1...|q|)\}, \frac{d}{dt} \left(\frac{\partial L}{\partial \dot{q_i}}\right)
 - \frac{\partial L}{\partial q_i}=Q
\end{equation}
Using just this equation, the modeling process is reduced to
specifying the kinetic and potential energy in the system.  Part of
the power of the method comes from the fact that this can be done
using Cartesian, polar, spherical, or any other generalized
coordinates.  Compared to the classic Newtonian, force-vector
based methods, the Euler-Lagrange equation is often a more direct
specification of the dynamics.

The Lagrangian modeling process consists of four steps:
\begin{enumerate}
\item Start with a description of the components of the system,
  consisting of rigid bodies and joints.  This description generally
  comes with a set of variable names which are collectively called
  called the generalized coordinates vector $q$.  Intuitively, each
  variable represents a quantity corresponding to one of the degrees
  of freedom for the system.  Usually, all of this information can be
  captured in an intuitive way in a {\em free body diagram}.
\item Determine the expression for the total kinetic and potential
  energy $T$ and $V$, respectively, of the system, in terms of the
  selected set of generalized coordinates $q$.
\item Identify and include any ``external forces'' $Q$ such as
  friction.\footnote{For this paper, we do not consider any such
    forces.}
\item Substitute the values into the Euler-Lagrange equation
  (\ref{eq-el}) for the variables of second the derivative of $q$.
\end{enumerate}

\begin{figure}
\centering
  \subfloat[Single]{\label{fig:single}\includegraphics[scale=0.4]{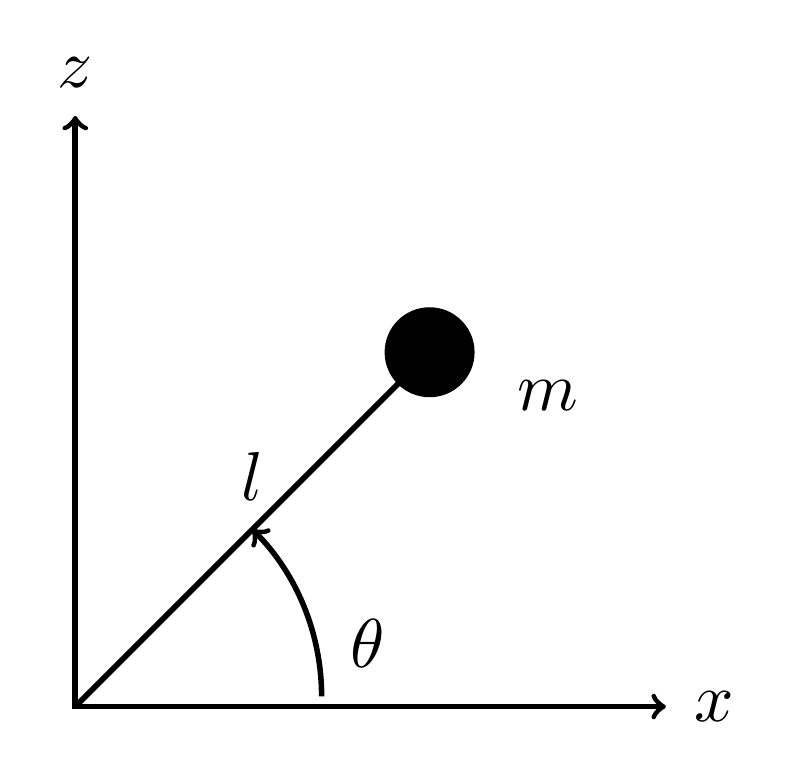}}
  \subfloat[Double]{\label{fig:double}\includegraphics[scale=0.2]{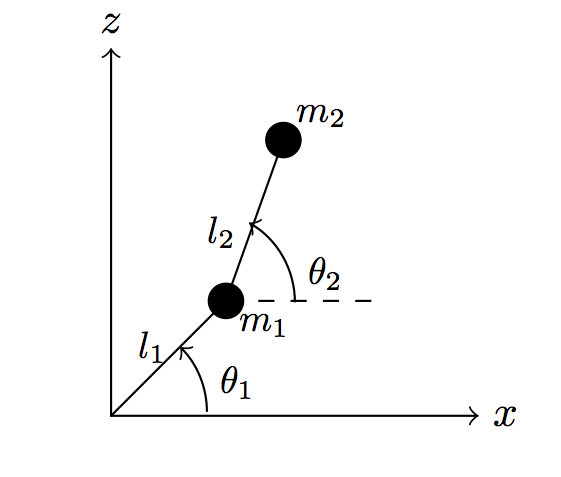}}
%\subfloat[three]{\label{fig:three}\includegraphics[scale=0.25]{DoublePendulum.jpg}}
\qquad  
\caption{Free body diagram for single and double pendulums}
\label{pendulums}
\end{figure}
   
\/*
\begin{figure}
  \centering
  \includegraphics[width=0.5\columnwidth]{single_pendulum.jpg}
  \caption{Free body diagram for a pendulum}\label{fig:simp-scr}
\end{figure}
*/

This process and its benefits can be illustrated with two small
examples.  The benefits apply whether or not the language supports
directed or undirected equations.  Figure~\ref{pendulums} presents a
free body diagram marked up with generalized coordinates ($\theta$ in
one, and $\theta_1, \theta_2$ in the other) for a single and a double
pendulum system.  First, we consider the single pendulum.  A direct
application of the angular part of Newton's law gives us the following
equation:
\begin{equation}
\ddot{\theta}=\frac{g}{l}\cos{\theta}
\end{equation}
which is easily expressed in Acumen as follows:
\begin{verbatim}
class pendulum (l)
private
  theta := 0;theta':=0;theta'':=0;
  g:=9.81;
end
  theta'' = g/l*cos(theta); 
end
\end{verbatim}

\/*
Newton Method
\begin{equation}
\sum{M}=I\ddot{\theta}=ml^2\ddot{\theta}=mgl\cos\theta
\end{equation}
Lagrange Method
\begin{equation}
L=T-V
\end{equation}
\begin{equation}
T=\frac{1}{2} mv^2=\frac{1}{2}m(\dot{\theta})^2; V=-mg\sin{\theta};
\end{equation}
\begin{equation}
\forall i\in\{1...|q|\}, \frac{d}{dt} \left(\frac{\partial L}{\partial \dot{q_i}}\right) 
- \frac{\partial L}{\partial q_i}=Q

\end{equation}
\begin{equation}
ml^2\ddot{\theta}=mgl\cos{\theta}
\end{equation}

Of note is how straightforward this model is. It requires no obvious
coordinate transforms (keen observer should note that this is
transitioning between a global coordinate frame and a body fixed
coordinate frame after a rotation about the global Y axis).  Also, at
this stage, the difference in complexity between Newton and Lagrangian
methods is minimal.  */

Lagrangian modeling can be applied to the single pendulum problem, but
Newton's method works well enough here.  However, Lagrangian modeling
does pay off for a double pendulum.  It is instructive for language
designers to recognize that such a seemingly small change in the
complexity of the rigid body makes the model most of us learn about in
high-school much more cumbersome than necessary.  Whether or not this
difficulty in modeling is due to weakness in Newtonian modeling or
intrinsic complexity in this seemingly simple example is not obvious:
The double pendulum is sophisticated enough to be widely used to model
a human standing or walking~\cite{pekarek2007discrete}, or a basic
two-link robot such as the MIT-manus~\cite{hogan1992manus}.

To derive a model for the double pendulum using Lagrangian modeling,
we proceed as follows:
\begin{enumerate}
\item We take $q=(\theta_1, \theta_2)$. Here, because the
  Euler-Lagrange equation is parameterized by a generalized coordinate
  vector, we could have chosen to use Cartesian coordinates $(x,z)$
  for each of the two points.  Here we chose angles because they
  resemble what can be naturally measured and actuated at joints.
\item The kinetic and potential energies are defined as follows:
\begin{equation}
T=\frac{1}{2} m_{1}v_{1}^2+\frac{1}{2}m_{2}v_{2}^2
\end{equation}
\begin{equation}
V=m_{1}gz_{1}+m_{2}gz_{2}
\end{equation}
where we have introduced shorthands for speeds $v_1^2=l_1^2 \dot{\theta_1}^2$ and
$v_2^2=v_1^2+\frac{1}{2}m_2(l_{1}^2\dot{\theta_{1}}^2+l_{2}^2\dot{\theta_{2}}^2+2l_{1}l_{2}\dot{\theta_{1}}\dot{\theta_{2}}\cos({\theta_{2}-\theta_{1}}))$, $z_1=l_1sin\theta_1$ and heights $z_2=z_1+l_2sin\theta_2$.  Substituting these terms we get:
\begin{multline} T=\frac{1}{2}m_{1}(l_1 \dot{\theta_{1}})^2
  \\+\frac{1}{2}m_2(l_{1}^2\dot{\theta_{1}}^2+l_{2}^2\dot{\theta_{2}}^2+2l_{1}l_{2}\dot{\theta_{1}}\dot{\theta_{2}}\cos({\theta_{2}-\theta_{1}}))
\end{multline}
\begin{equation}
V=m_{1}gl_{1}\sin{\theta_{1}}+m_{2}gl_{2}\sin{\theta_{2}}+m_{2}gl_{1}\sin{\theta_{1}};
\end{equation}
\item We assume frictionless joints, and so there are no
  external forces ($Q=0$).
\item By substitution and (manual) symbolic differentiation we get:
\begin{multline}
(m_{1}+m_{2})l_{1}^2\ddot{\theta_{1}}+m_{2}l_{1}l_{2}\ddot{\theta_{2}}\cos{(\theta_{1}-\theta_{2})}\\+m_{2}l_{1}l_{2}\dot{\theta_{1}}^2\sin({\theta_{1}-\theta_{2}})+m_{1}m_{2}gl_{1}\cos{\theta_{1}}=0
\end{multline} \begin{multline}
m_{2}l_{2}^2\ddot{\theta_{2}}+m_{2}l_{1}l_{2}\ddot{\theta_{1}}\cos{(\theta_{1}-\theta_{2})}-\\m_{2}l_{1}l_{2}\dot{\theta_{1}}^2\sin{(\theta_{1}-\theta_{2})}+m_{2}gl_{2}\cos{\theta_{2}}=0
\end{multline} 
It is important to note in this case that, while these are ordinary
differential equations (ODEs), they are not in the explicit form
$X'=E$.  Rather, they are in implicit form, because the variable we
are solving for ($X'$) is not alone on one side of the equation.
Using Gaussian elimination (under the assumption that the masses and
length are strictly greater than zero), we get: \begin{multline}
  \ddot{\theta_{1}}=m_{2}l_{2}(\ddot{\theta_{2}}\cos(\theta_1-\theta_2)+\dot\theta_{2}^2\sin(\theta_1-\theta_2)
  \\+(m_{1}+m_{2})g\cos(\theta_{1}))/(-l_{1}(m_{1}+m_{2})) \end{multline}
\begin{multline}
\ddot{\theta_{2}}=m_{2}l_{1}(\ddot{\theta_{1}}\cos(\theta_1-\theta_2)-\dot\theta_{1}^2\sin(\theta_1-\theta_2)
\\+m_{2}g\cos(\theta_{2}))/-l_{1}m_{2} \end{multline}
\end{enumerate}
The following code shows these equations expressed in Acumen:

\/*
\begin{figure}
  \centering
  \includegraphics[width=0.6\columnwidth]{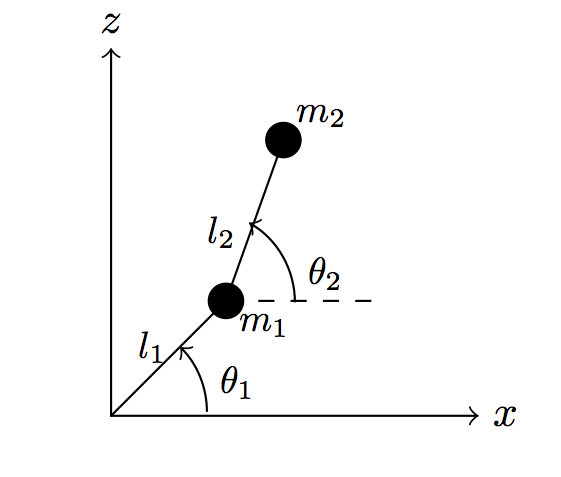}
  \caption{Free body diagram of a double pendulum}\label{fig:doub-scr}
\end{figure}
*/

\begin{verbatim}
class double_pendulum (m_1, m_2, L_1, L_2)
private 
  t_1 := 0; t_2 := 0; 
  t_1' := 0; t_2' := 0;
  t_1'' := 0; t_2'' := 0;
  g:=9.81;
end
  t_1'' = 
   (m_2*L_2*(t_2''*cos(t_1-t_2)
             +t_2'^2*sin(t_1-t_2))
    + (m_1+m_2)*g*cos(t_1))
   *(-1)/((m_1+m_2)*L_1);
  t_2'' = 
   (m_2*L_1*(t_1''*cos(t_1-t_2)
             -t_1'^2*sin(t_1-t_2))
    +m_2*g*cos(t_2))
   *(-1)/(m_2*L_2); 
end
\end{verbatim} 

\/*
\subsection{Dynamics of a 3D Rod}\label{sec-3Drod}
Now, as a more general, 3D problem, consider a rod which holds apart
two masses (of $m/2$ each) at a given distance (visualized
in~\Fig{rod-schem}). If there are force vectors $p$ and $q$ acting on
each end, what is the resulting acceleration on the system? First, one
must develop the equations for the dynamics.
\begin{figure}
  \centering
  \includegraphics[width=0.75\columnwidth]{3DRodTikZ.jpg}
  \caption{Free body diagram of a rod}\label{fig:rod-schem}
\end{figure}

\Fig{rod-schem} is the free body diagram that was used in steps 1 and
2 of the Lagrangian method to make the model illustrated
below. The original definition of the problem is of a rod, of length
$l$ with point masses $m/2$ at endpoints $p$ and $q$. We defined a
global coordinate system (the one used in the model, X, Y, Z) and a
body-fixed, non rotating frame ($x, y, z$) located at $c$, the center
of the rod, and one body-fixed frame with the $\hat{x}$ axis pointing
towards $q$ $(\hat{x}, \hat{y}, \hat{z})$. In the picture, $\hat{y}$
points into the page, and $y$ is also behind the
page. ~\Fig{rod-schem} also shows how the angles $\alpha$ and $\beta$
are defined. See Appendix \ref{app-rod} for the derivation behind the
following model of the dynamics of such a rod:

\begin{verbatim}
class rod (m,l,c0,a0,b0,D)
 private 
  c := c0; c' := [0,0,0]; c'' := [0,0,0];
  a := a0; a' :=0; a'' := 0;
  b := b0; b' :=0; b'' := 0;  g := 9.81;
  torque := [0,0,0]; fp := [0,0,0]; 
  fq := [0,0,0]; p := [0,0,0];   
  q := [0,0,0]; vp := [0,0,0];
  vq := [0,0,0]; end
 torque = 
     cross(fp,l/2*[cos(a)*cos(b),
                   sin(a)*cos(b),sin(a)]) -
     cross(fq,l/2*[cos(a)*cos(b),
                   sin(a)*cos(b),sin(a)]);
  c'' = (fp + fq)/m - g * [0,0,1];
  a'' = dot(torque,[0,1,0])/(m*l^2) - 
           1* b'^2 * cos(a)*sin(a);
  b'' = dot(torque,[0,0,1])/(m*l^2) -
          1 * b'*a'*cos(a)*sin(a)/cos(a)^2;
  p = c -l/2* [cos(a)*cos(b),
               cos(a)*sin(b),sin(a)];
  q = c +l/2* [cos(a)*cos(b),
               cos(a)*sin(b),sin(a)];
 end
\end{verbatim}
This seemingly simple example of a 3D rod is, in fact, sufficient to
illustrate key aspects of 3D rigid body dynamics, including nonlinear
kinematics, dynamics, and control, as well as under-actuation. It is
easy to combine this system with a controller that works to move the
point $p$ to a predetermined location, but a more interesting and
difficult goal would be to control the location of the point $q$
during the process, and see whether the controller can maintain
unstable positions, as shown in~\Fig{rod-scr}.  This problem provides
a natural starting point to study challenging questions such as the
control of an inverted pendulum in 3D.
\begin{figure}
  \centering
  \includegraphics[width=0.8\columnwidth]{rodanimationV2.jpg}
  \caption{Snapshots from a 3D visualization of the Rod and PID classes allow the
    user to observe the rigid body dynamics, and the effect of the
    controller on overshoot and the elimination of steady-state
    error.}
\label{fig:rod-scr}
\end{figure}
*/

\begin{figure}
\centering
\includegraphics[width=0.8\columnwidth]{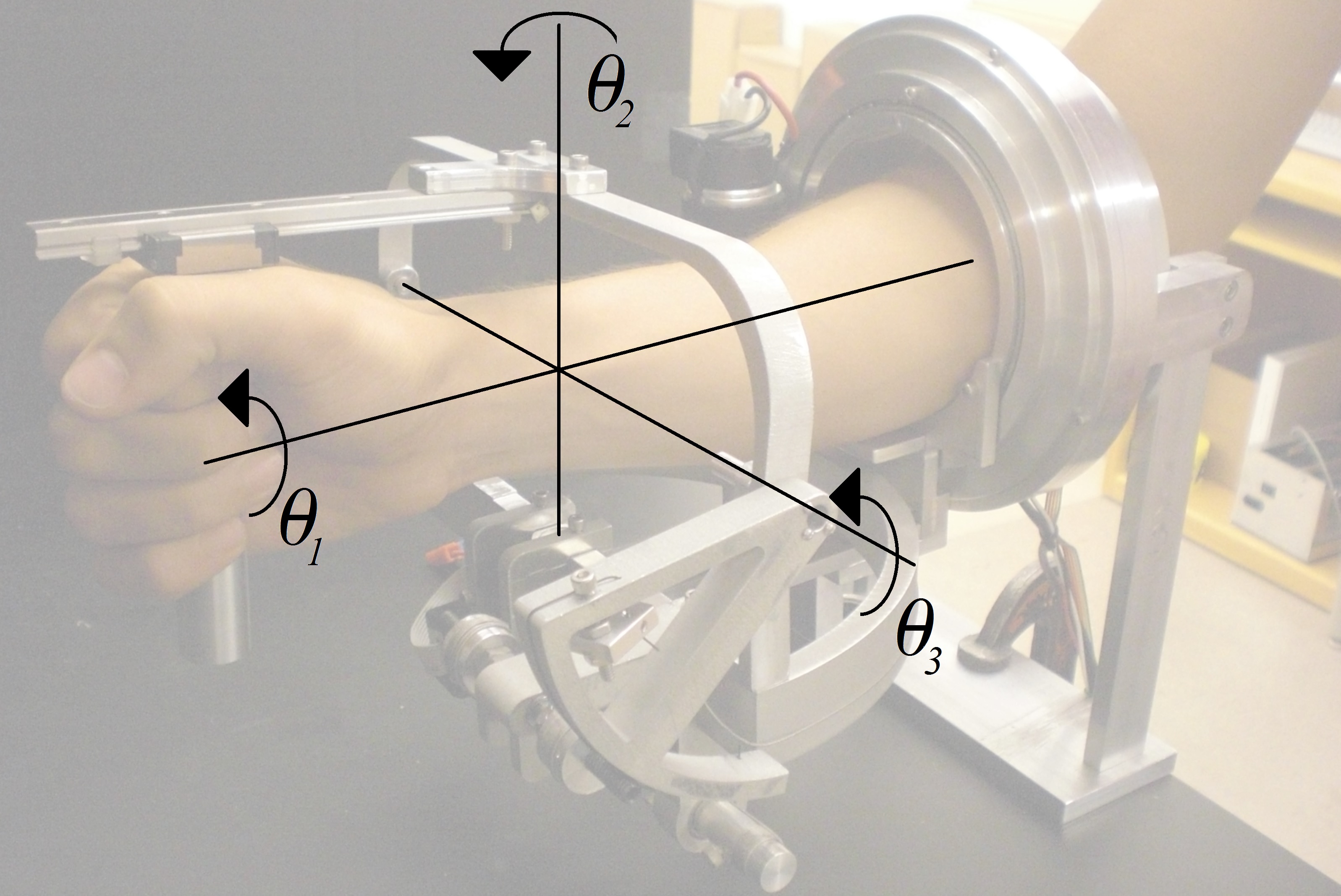}
\caption{The RiceWrist-S, with superimposed axes of rotation}
\label{fig-RWSAxes}
\end{figure}

\section{The RiceWrist-S Robot}\label{sec-RWS} 
Equipped with an understanding of Lagrangian modeling in the manner
presented above, engineers model multi-link robots much more directly
than with the Newtonian method.  In this section we present one such
case study using the RiceWrist-S research Robot.

\subsection{Background}
With an increasing number of individuals surviving once fatal injuries, 
the need for rehabilitation of damaged limbs is growing rapidly. Each
year, approximately 795,000 people suffer a stroke in the United
States, where stroke injuries are the leading cause of long-term
disability.
\/* incurring an annual estimated cost of \$38.6 billion.  [Heart
  Disease and Stroke Statistics 2013 Update (Chad: See ICORR2013
  submission)].Furthermore, there are approximately 12,000 incidences
of Spinal Cord Injury (SCI) each year [Spinal Cord injury facts and
  figures (Chad)] with an estimated total yearly direct and indirect
costs of \$14.5 billion and \$5.5 billion, respectively [Berkowitz,
  Spinal cord injury... (Chad)]. */
The RiceWrist-S~\cite{RiceWrist} is an exoskeleton robot designed to
assist in the rehabilitation of the wrist and forearm of stroke or
spinal cord injury patients (Fig.~\ref{fig-RWSAxes}).  It consists of
a revolute joint for each of the three degrees of freedom at the
wrist.  Because it has three rotational axes intersecting at one
point, a good starting point to modeling it is the gimbal, a commonly
studied mechanical device that also features several rotational axes
intersecting at one point.
\subsection{Analytical Model}

\begin{figure} \centering \includegraphics[width=0.8\columnwidth]{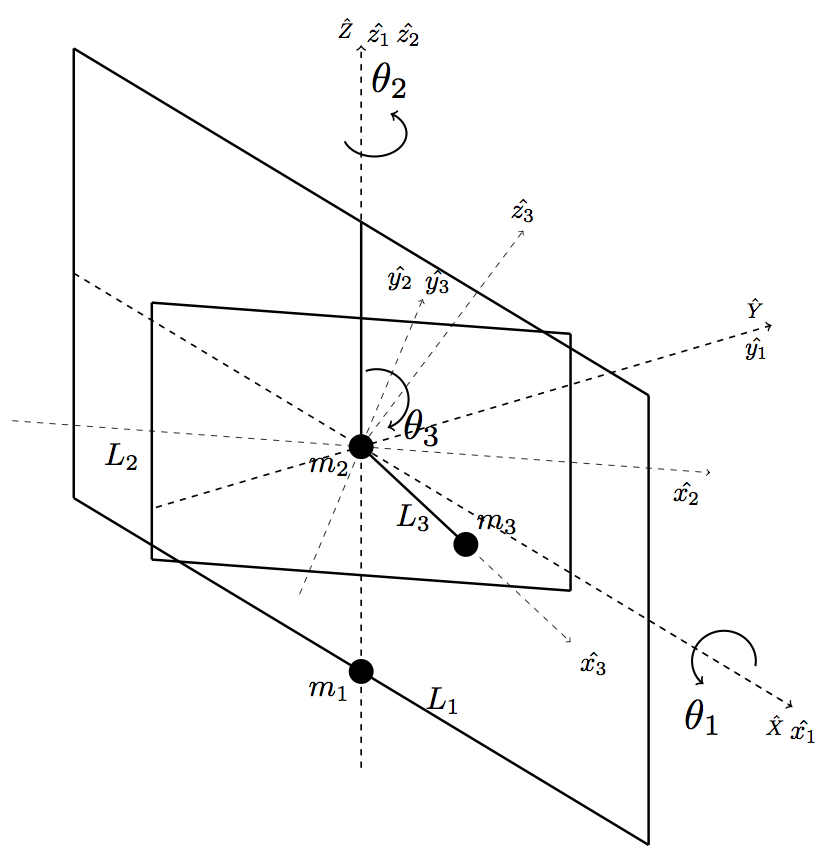}
\caption{Free body diagram of the RiceWrist-S as a gimbal}\label{fig:RWS-scr} \end{figure}
We can apply the Lagrangian modeling process to determine the dynamics
of a gimbal as follows:
\/* but, due to page limits, the model can be found online~\cite{Comparison}*/

\begin{figure}
\centering \subfloat[Rotation about $\theta_1$]{\label{fig:RWS1}\includegraphics[scale=0.2]{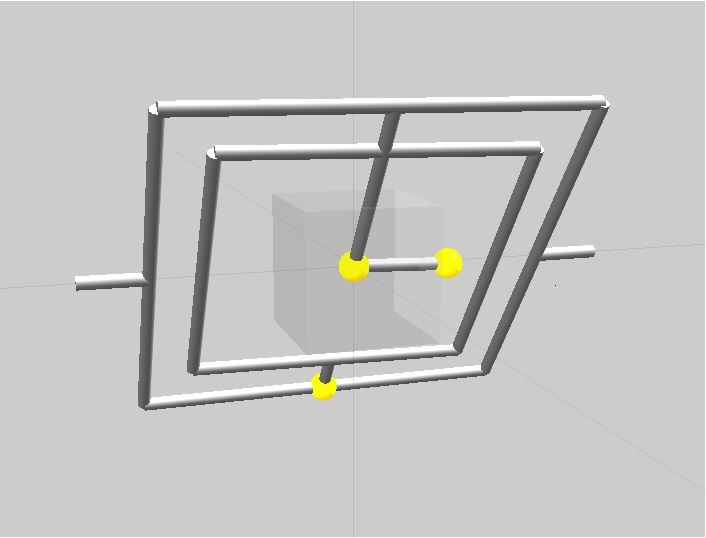}} \subfloat[Rotation about $\theta_2$ ]{\label{fig:RWS2}\includegraphics[scale=0.2]{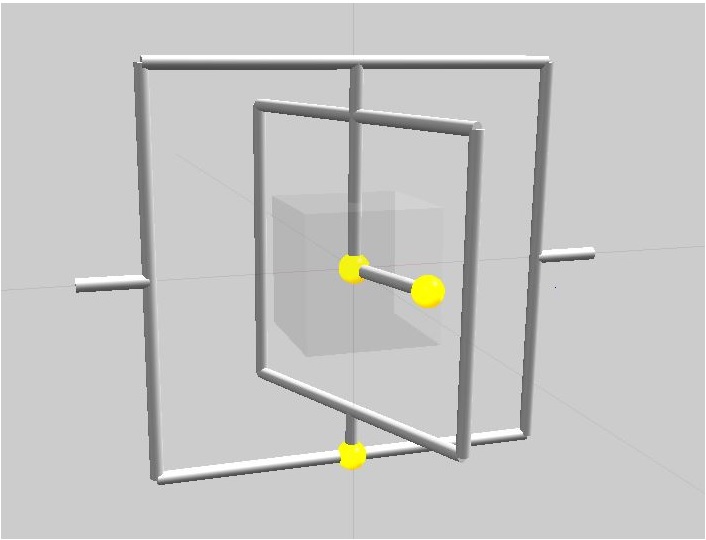}}
\qquad  \subfloat[Rotation about $\theta_3$]{\label{fig:RWS3}\includegraphics[scale=0.2]{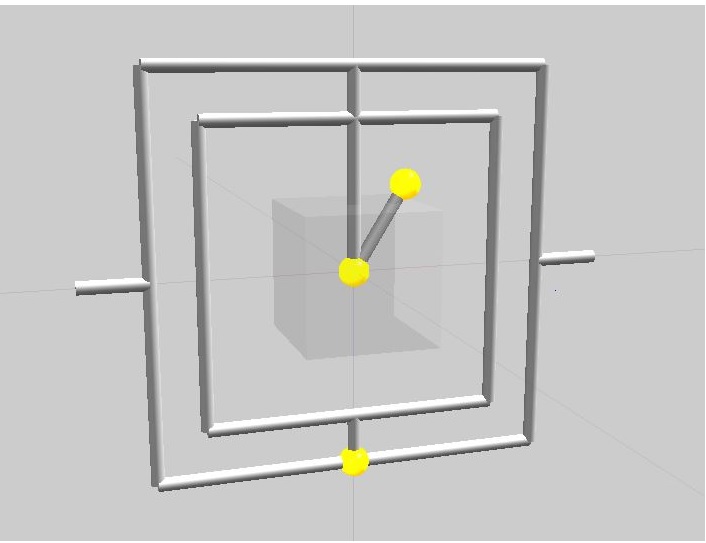}}
\subfloat[Compound motion]{\label{fig:RWS4}\includegraphics[scale=0.2]{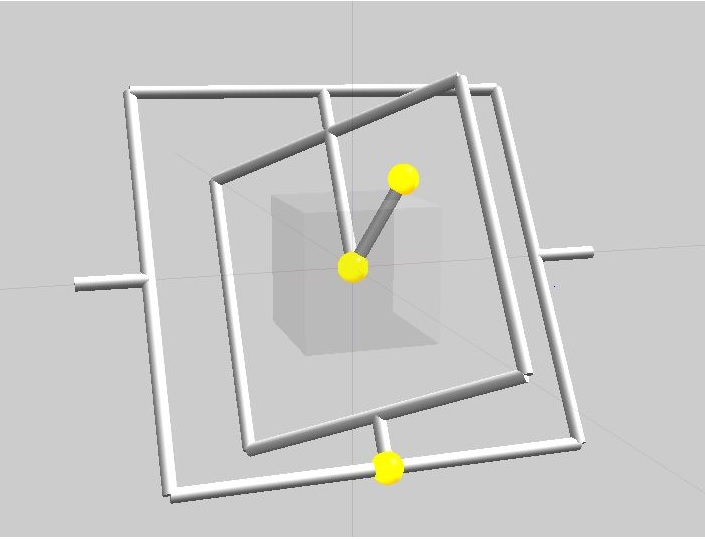}}
\caption{RiceWrist-S modeled as Gimbal in Acumen.} \label{RWS-screenshots}
\end{figure}

\begin{enumerate}
\item We take $q=(\theta_1,\theta_2,\theta_3)$, where each of the
  angles corresponds to one of the three rotations possible in the
  RiceWrist-S (Fig.~\ref{fig:RWS-scr}). We chose to represent the mass
  of the system as centralized to three locations, one at the
  origin, one at the bottom of the outermost ring, and one at the end
  of the third link.  The masses in this figure correspond to the
  motors and handle depicted in Fig.~\ref{fig-RWSAxes}.
\item To describe the energies concisely, it is convenient to use the
  following angular velocities of the gimbal frames in the kinetic
  energy terms, and the resulting heights for the potential energy
  terms:
    \begin{align}
\begin{split}\label{eq:21}
    \omega_1 =\,& \dot{\theta_1} \cdot \hat{x_1}\\
        \end{split}\\
\begin{split}
    \omega_2  =\,&\dot{\theta_1}\cdot \hat{x_1} + \dot{\theta_2}\cdot \hat{z_2}\\ 
            \end{split}\\
            \begin{split}
    \omega_3 =\,&\dot{\theta_1}\cdot \hat{x_1} + \dot{\theta_2}\cdot \hat{z_2}+\dot{\theta_3}\cdot \hat{y_3}\\
    \end{split}
\end{align}
where $\hat{x_i}\hat{y_i}\hat{z_i}$ refers to the unit vector and coordinate frame about which these rotations occur, as shown in Fig.~\ref{fig:RWS-scr}.
Here, the $\omega_i$ terms correspond to the $m_i$ masses, and describe the angular velocities of that mass. Since this is a complex rotational system, many of the rotations do not occur in the coordinate frames of the respective gimbal. Therefore, in order to express each $\omega_i$ in terms of the same coordinate frame, we applied the following coordinate transforms: 
%\begin{multline}
%\omega_1=\dot{\theta_1} \cdot \hat{x_1}\\ 
%\omega_2 =\dot{\theta_1}\cdot \hat{x_1} + \dot{\theta_2}\cdot \hat{z_2}\\ 
%\omega_3=\dot{\theta_1}\cdot \hat{x_1} + \dot{\theta_2}\cdot \hat{z_2}+\dot{\theta_3}\cdot \hat{y_3}
%\end{multline}
\begin{align}
\begin{split}\label{eq:21}
    \omega_1 =\,& \dot{\theta_1} \cdot \hat{x_1}\\
        \end{split}\\
\begin{split}
    \omega_2  =\,&\dot{\theta_1}(\cos(\theta_2)\cdot \hat{x_2} -\sin(\theta_2)\cdot \hat{y_2})+  \dot{\theta_2}\cdot \hat{z_2}\\
            \end{split}\\
            \begin{split}
    \omega_3 =\,&(\dot{\theta_1}(\cos(\theta_2)\cos(\theta_3))+\dot{\theta_2}(-\sin(\theta_3)))\cdot \hat{x_3} \\
    &+ (-\dot{\theta_1}\sin(\theta_2)+\dot{\theta_3})\cdot \hat{y_3} +  (-\dot{\theta_1}\sin(\theta_3)\cos(\theta_2)\\
    &-\dot{\theta_2}\cos(\theta_3))\cdot \hat{z_3}\\
        \end{split}
\end{align}

Next, we express the heights above the predefined plane of zero
potential energy (in Fig.~\ref{fig:RWS-scr}, the $XY$ plane) of each
of the masses $m_1, m_2, m_3$, respectively, as the following:
\begin{align}
  \begin{split}
    h_1 =\,& l_2\cos(\theta_1)\\
   \end{split}\\
   \begin{split} 
    h_2 =\,& 0\\
    \end{split}\\
   \begin{split}
     h_3 =\,& l_3\sin(\theta_1)\sin(\theta_3)\\
    \end{split}
\end{align}

%\begin{multline}
%\omega_1=\dot{\theta_1} \cdot \hat{x_1}\\
%\omega_2 =\dot{\theta_1}(\cos(\theta_2)\cdot \hat{x_2} -\sin(\theta_2)\cdot \hat{y_2})+  \dot{\theta_2}\cdot \hat{z_2}\\
%\omega_3 =(\dot{\theta_1}(\cos(\theta_2)\cos(\theta_3))+\dot{\theta_2}(-\sin(\theta_3)))\cdot \hat{x_3} +\\ (-\dot{\theta_1}\sin(\theta_2)+\dot{\theta_3})\cdot \hat{y_3} + \\ (-\theta_1\sin(\theta_3)\cos(\theta_2)-\dot{\theta_2}\cos(\theta_3))\cdot \hat{z_3}
%\end{multline}
%With this completed, the $T$ and $V$ terms can be easily defined, as shown below. 
%\begin{equation}
%T=\frac{1}{2}(I_1\omega_1^2+I_2\omega_2^2+I_3\omega_3^2)
%\end{equation}

With this completed, the $T$ and $V$ terms can be quickly and easily
defined. Since this is a rotational only system, $T$ is defined as the
sum of the rotational energy terms, shown below:
\begin{equation}
T=\frac{1}{2}(I_1\omega_1\cdot\ \omega_1+I_2\omega_2\cdot \omega_2+I_3\omega_3\cdot \omega_3)
\end{equation}
Where $I_i$ is the rotational inertia corresponding to $\theta_i$, and
$\omega_i$ is defined as above.  And since there are no potential
energy storage elements other than those caused by gravity, $V$ can be
expressed with these heights:

\begin{multline}
V =m_1gh_1+m_2gh_2+m_3gh_3\\
   = -m_1gl_2\cos(\theta_1) + m_3gl_3\sin(\theta_1)\sin(\theta_3)
\end{multline}

\item Again, we assumed frictionless joints, so $Q=0$
\item After substitution and (manual) symbolic differentiation, we get an implicit set of
  equations.  We solve these for $q''$ to get the equations
  of the systems dynamics.  
\/*  
%\begin{multline}
%\omega_1=\dot{\theta_1} \cdot \hat{x_1}\\
%\omega_2 =\dot{\theta_1}(\cos(\theta_2)\cdot \hat{x_2} -\sin(\theta_2)\cdot \hat{y_2})+  \dot{\theta_2}\cdot \hat{z_2}\\
%\omega_3 =(\dot{\theta_1}(\cos(\theta_2)\cos(\theta_3))+\dot{\theta_2}(-\sin(\theta_3)))\cdot \hat{x_3} +\\ (-\dot{\theta_1}\sin(\theta_2)+\dot{\theta_3})\cdot \hat{y_3} + \\ (-\theta_1\sin(\theta_3)\cos(\theta_2)-\dot{\theta_2}\cos(\theta_3))\cdot \hat{z_3}
%\end{multline}

The $\omega_1, \omega_2, \omega_3$ terms must be defined in the coordinate frames of their respective gimbal frame (i.e. $\hat{x_1}...$ corresponds to the outermost gimbal frame, $\hat{x_2}...$ corresponds to the inner gimbal frame, and $\hat{x_3}...$ corresponds to the inner link) as shown below.
\begin{align}
\begin{split}\label{eq:21}
    \omega_1 =\,& \dot{\theta_1} \cdot \hat{x_1}\\
        \end{split}\\
\begin{split}
    \omega_2  =\,&\dot{\theta_1}(\cos(\theta_2)\cdot \hat{x_2} -\sin(\theta_2)\cdot \hat{y_2})+  \dot{\theta_2}\cdot \hat{z_2}\\
            \end{split}\\
            \begin{split}
    \omega_3 =\,&(\dot{\theta_1}(\cos(\theta_2)\cos(\theta_3))+\dot{\theta_2}(-\sin(\theta_3)))\cdot \hat{x_3} \\
    &+ (-\dot{\theta_1}\sin(\theta_2)+\dot{\theta_3})\cdot \hat{y_3} +  (-\dot{\theta_1}\sin(\theta_3)\cos(\theta_2)\\
    &-\dot{\theta_2}\cos(\theta_3))\cdot \hat{z_3}\\
        \end{split}
\end{align}

\begin{equation}
V=m_1gh_1+m_2gh_2+m_3gh_3
\end{equation} 
Where $h_1, h_2, h_3$ are the heights above the predefined plane of zero potential energy for $m_1, m_2, m_3$, respectively. Using kinematic relationships to solve for these heights:
\begin{align}
  \begin{split}
    h_1 =\,& l_2\cos(\theta_1)\\
   \end{split}\\
   \begin{split} 
    h_2 =\,& 0\\
    \end{split}\\
   \begin{split}
     h_3 =\,& l_3\sin(\theta_1)\sin(\theta_3)\\
    \end{split}
\end{align}

\begin{equation}
V= -m_1gl_2\cos(\theta_1) + m_3gl_3\sin(\theta_1)\sin(\theta_3)
\end{equation}
*/
%Lagrangian equation:
\/*
\begin{equation}
\frac{d}{dt} \left(\frac{\partial L}{\partial \dot{\theta_{1}}}\right) - \left(\frac{\partial L}{\partial {\theta_{1}}}\right) = 0
\end{equation}
Expands to:
*/
\begin{multline}
\frac{1}{2}\frac{d}{dt} \left(I_1 \frac{\partial \omega_1\cdot \omega_1}{\partial \dot{\theta_{1}}}+ I_2 \frac{\partial \omega_2\cdot \omega_2 }{\partial \dot{\theta_{1}}}+ I_3\frac{\partial \omega_3\cdot \omega_3 }{\partial \dot{\theta_{1}}}\right) +\\m_1 g l_2 sin(\theta_1) +m_3 g l_3 cos(\theta_1) sin(\theta_3) =0
\end{multline}
\/*
\begin{equation}
\frac{d}{dt} \left(\frac{\partial L}{\partial \dot{\theta_{2}}}\right) -  \left(\frac{\partial L}{\partial {\theta_{2}}}\right) = 0
\end{equation}
Expands to:
*/
\begin{multline}
\frac{1}{2}\frac{d}{dt} \left(I_2\frac{\partial \omega_2\cdot \omega_2 }{\partial \dot{\theta_{2}}}+I_3\frac{\partial \omega_3\cdot \omega_3 }{\partial \dot{\theta_{2}}}\right) \\- 
\frac{1}{2} \left(I_2 \frac{\partial \omega_2\cdot \omega_2 }{\partial {\theta_{2}}}+I_3 \frac{\partial \omega_3\cdot \omega_3}{\partial {\theta_{2}}}\right) = 0
\end{multline}
\/*
\begin{equation}
\frac{d}{dt} \left(\frac{\partial L}{\partial \dot{\theta_{3}}}\right) -  \left(\frac{\partial L}{\partial {\theta_{3}}}\right) =  0
\end{equation}
Expands to:
*/
\begin{multline}
\frac{1}{2}\frac{d}{dt} \left(I_3\frac{\partial \omega_3\cdot \omega_3}{\partial \dot{\theta_{3}}}\right) - \frac{1}{2} \left(I_3 \frac{\partial \omega_3\cdot \omega_3}{\partial {\theta_{3}}}\right) \\-m_3 g l_3 sin(\theta_1)cos(\theta_3) = 0
\end{multline}

% \frac{\partial L}{\partial q_i}=Q
%\end{equation} 

\end{enumerate}
After computing the static partial derivatives, and the time
derivatives, the resulting ODE's are again, easy to express in Acumen.
But they are too long to include here, as is the code~\cite{RW-S}.
What is significant about this situation is that these equations are
linear in $\ddot{q}$ even though the system is also a non-linear
differential equation (when we consider $q$, $\dot{q}$ and
$\ddot{q}$).  Because the system is linear in $\ddot{q}$, we can use
standard Gaussian elimination to solve for $\ddot{q}$.  This converts
these implicit equations into an explicit form that is readily
expressiable in Acumen. Fig.~\ref{RWS-screenshots} shows the compound
motion of the RW-S Gimbal model.
 \/*

\subsection{Compass Gait} Beyond the need for stroke and SCI rehabilitation,
many individuals suffer a traumatic injury that requires the
replacement of an amputated limb. This need to recreate human-like
motion falls neatly into goals set forth by the robotics community,
not only for advanced prosthetics [add some of Ames’ citations here?],
but for robotic motion in general. Recreating this advanced type of
gait would allow for robots to handle the complex terrain and slopes
that people can easily navigate that frustrate even advanced robots
[cite Ames paper]. The model presented here, referred to as a Compass
Gait model, is simple enough to lend itself to study but still
possesses many of the characteristics of human walking, making it a
favorable place to being studying gait patterns, and optimizing
control that could be used for bipedal humanoid robots as well as
powered prostheses [Pekarek, Ames, Marsden, 2007].

Reference 2007 paper by Pekarek, Ames, and Marsden that says that “The
compass gait biped is a two-dimensional bipedal robot that is simple
enough to be amendable [sic] to analysis, yet complex enough [to]
display a wealth of interesting phenomena.” The compass gait model is
a great place to start studying gait patterns, and optimizing control

PLACEHOLDER FOR COMPARISON TO DOUBLE PENDULUM (DISCUSSION OF ADDED
DISCRETE EVENTS)

\begin{figure}
  \centering
  \includegraphics[width=1\columnwidth]{CompassGaitanimation.jpg}
  \caption{Snapshots from a 3D visualization of the Passive Compass Gait class allow the
    user to observe the dynamics of the model as it walks down a slope.}
\label{fig:Compass-scr}
\end{figure}
*/

\section{Discussion}\label{sec-Discussion}

In this section, we step back to reflect on how to interpret this
experience with expressing these various sytems in the small core
language described at the outset of the paper.

\subsection{The Role of a Core Language}

At the highest level, Acumen can be seen as an untyped core formalism
for a subset of hybrid systems modeling and simulation languages such
as Modelica~\cite{Modelica}, Scicos~\cite{Nikoukhah-2006} and
{Simcape}~\cite{Simscape}.  It is easy to see that the CPS aspects
that Acumen can express should be expressible by ``larger'' languages,
but it is less clear what finding a weakspot in the expressivity of
Acumen means for larger languages. In reasoning formally about
expressivity limitations, it is beneficial that Acumen is untyped: any
static type discpline adds restrictions, and whether or not such
restrictions affects expressivity complicates such reasoning.  While
removing typing avoids this particular difficulty, it does not remove
all the difficulties.  Eventually, we expect to add a typing discpline
to Acumen, but no particular direction has been decided at this time.

With several annual workshops related to its design and applications,
Modelica is the most actively studied of the hybrid languages.  It is
a large language with many constructs.  Acumen is small, with only a
few features that are specific to hybrid systems. For example, Open
Modelica's parser~\cite{parser} is roughly four times longer than the
parser for Acumen.  If Acumen can be seen as a subset of Modelica, it
is reasonable to assume that the CPS aspects that Acumen can express
can also be expressed in Modelica.  At the same time, if features of
CPS models are not easily expressed in Acumen, it is not immediately
obvious wether or not they can be expressed in Modelica.

Another point that requires care in making the connection is the
relation between Acumen's notion of an object class and the
abstraction mechanisms found in other languages.  So far, our
comparisons focus on issues relating to the expression and statement
parts of these languages.

\subsection{Supporting Lagrangian Modeling}

The experience with the RiceWrist-S Robot suggests the need for
Lagrangian modeling, which in turn points out to the need for two
language features.  The first feature is {\em static} partial
derivatives.  It has been observed elsewhere that the type of partial
derivatives used in the Euler-Lagrange equation for rigid body
dynamics can be removed at compile time (or ``statically'')
\cite{zhu2010mathematical}.  The pendulum examples show that CPS
modeling languages should support the definition of explicit
Langrangian models (requiring partial derivatives) because it
simplifies the task of modeling rigid body systems, reducing the
amount analysis, description, and algebraic manipulation the user
would have to carry out.

Support for static partial derivatives in larger modeling and
simulation tools seems sparse.  While the Modelica standard (at least
until recently) did not support partial derivatives, one
implementation of Modelica, Dymola \cite{OlssonUad}, does provide this
support.  Static partial derivatives are clearly an important feature
to include in any DSL aiming to support CPS design.

In response to this observation, we have already introduced support
for static partial derivatives in Acumen.  The second language feature
required (but is not sufficient without the first) to support
Lagrangian modeling is implicit equations.  The utility of implicit
equations can be seen by reviewing the double pendulum example.
Implicit equations are present in Modelica and Simscape.  The
construct does increase the reusability and flexibility of models,
making composite models more concise.  At this point it is not
entirely clear to us how to best introduce implicit equations into a
core language.  To address this issue, we are currently implementing
and experimenting with several different approaches from the
literature.  While different approaches may be better suited for
different purposes (such as numerical precision, runtime performance,
or others), our purpose behind this activity is to better understand
how to position implicit equations in the context of other CPS DSL
design decisions.  A particularly interesting question here is whether
this feature must be a primitive in the language or whether it can be
treated as a conservative extension~\cite{Felleisen90onthe} of a core
that does not include it.

\/*
To better understand the relation between the two languages, we
conducted a side-by-side comparison by translating
\begin{itemize}
\item the models studied in this paper into Modelica,
\item the examples in a recent Modelica textbook~\cite{fritzson2011introduction} into Acumen
\end{itemize}
The full side-by-side comparison is available
online~\cite{Comparison}.  Most of the observations from this
comparison can illustrated using the small example presented in Table
\ref{tb:tablename}.
\begin{figure*}\label{code}
%\begin{table*}\label{tableone}
\begin{tabular}{p{8.5cm} p{8.5cm}}
\begin{verbatim}
// Modelica model
class mass_1d
 Real m,p,f,pp;
 equation
  der(p) = pp;
  der(pp) = f / m;
end mass_1d;

class example_1BB
  mass_1d m(m = 10, 
            p(start = 3), 
            pp(start = 0));
equation
  m.f = m.m * (-9.8);
  when m.p < 0 and m.pp < 0 then
      reinit(m.pp, -0.9 * pre(m.pp));
  end when;
end example_1BB;
\end{verbatim} 
&
\begin{verbatim}
// Core language model
class mass_1d (m,p0)
 private p:=p0; p':=0; p'':=0; f:=0 end

 p'' = f/m;

end

class example_1BB ()
 private m := create mass_1d (10,3) end


 m.f = m.m * -9.8;

 if (m.p<0 && m.p'<0)
  m.p' := -0.9 * m.p'
 end
end
\end{verbatim} 

\end{tabular}
% \hline
\caption{A bouncing ball model Modelica and Acumen}
\label{tb:tablename}

%\end{table*}
\end{figure*}

The key observations are as follows:
\begin{itemize}
\item Models expressed in the proposed core language are generally
  slightly shorter than those in Modelica.  This is primarily due to
  more concise notation for key operations such as discrete assignment
  (or reinitialization in Modelica terminology), parameter passing.
  However, part of this is also due to Acumen being untyped.
\item In some instances, Modelica has restrictions that have no
  obvious motivation. For example, Modelica only supports first
  derivatives of variables (written \verb|der(x)| for a variable
  \verb|x|) and not higher-order derivatives.  The only recourse is to
  declare a new variable and connect them together (essentially
  performing manual order reduction on the differential equation).
  The proposed core language avoids such restrictions.  Acumen tries
  to avoid such restrictions.
\item The syntax for hybrid (continuous/discrete) systems in Modelica
  appears to be more complex than necessary.  In the example shown in
  Table \ref{tb:tablename}, the core language uses one
  \verb|if|-statement and a discrete assignment (\verb|:=|) to
  describe the bouncing event of a falling ball, whereas Modelica's
  syntax requires the use of multiple key words such as \verb|when|,
  \verb|reinit| and \verb|pre| to express the same operation.
\end{itemize}
*/

\section{Conclusions}\label{sec-discussion} 

This paper reports our most recent findings as we pursue an
appropriate core language for CPS modeling and simulation.  In
particular, we present two examples.  Our first example, a quadcopter,
is significantly more sophisticated than any system considered in the
first part.  It shows that the core language proposed in Part I can
express more sophisticated systems than was previously known.  The
second example, the RiceWrist-S, led us to understand the need for
Lagrangian modeling, and as a result, provides us with concrete
justification for introducing support for static partial derivatives.
The most important lesson from this case study was that a DSL aimed at
supporting CPS design should provide static partial derivatives and
implicit equations.  In order to ensure that we can push this line of
investigation further, we have already developed a prototype of the
former, and are working to realize the latter.

\/*including the Hamiltonian method,
Kane's method, the Denavit-Hartenberg convention, and Featherstone's
algorithm, each with their own specialized use in dynamics. Also,*/

\/* However, the Hamiltonian and Kane’s [from Gillespie paper: (Schaechter,
1988), (Kane and Levinson, 1999)-textbook, (Autolev, 2003)-www.autolev.com]
method are powerful methods based on the Lagrangian method. Kane’s method is
a more automated way for calculating the dynamics of a system, and is used in
the AUTOLEV solver (see Brent Gillespie’s paper again).  The Hamiltonian method
has far reaching applications including quantum mechanics applications. 

One particularly popular advancement, the Denavit-Hartenberg
convention [Denavit, Jacques; Hartenberg, Richard Scheunemann
  (1955). "A kinematic notation for lower-pair mechanisms based on
  matrices". Trans ASME J. Appl. Mech 23: 215–221.], is particularly
useful for deriving the kinematic equations describing robotic
mechanisms.

Also, Featherstone’s algorithm [Marcie look at proposal - for dynamic equation
derivation, has good properties for simulation. also Kane’s methods -- look at
Brent’s old paper and see if it describes the other methods nicely) particularly
powerful for roboticists [online copy of textbook which proposed algorithm].*/

\section*{Acknowledgements} We would like to thank the reviewers of DSLRob 2012
for valuable feedback on part I of this paper, and the reviewers of
DSLRob 2013 and Adam Duracz for comments and feedback on this draft.
%% \appendices

\bibliographystyle{unsrt}
\bibliography{local}
\hide{
\appendix

\section{Derivation of Equations for Rod Dynamics}\label{app-rod}

Using standard Lagrangian methods we can derive the model for the rod
as follows: The vector of generalized coordinates $q$ is comprised of
the independent degrees of freedom of the system.

\begin {equation}
q=[ x, y, z, \alpha, \beta ] 
\end{equation}
Define $L$ as the difference between the kinetic ($T$) and potential ($V$) energy terms:
\begin{equation}
L=T-V
\end{equation}
\begin{equation}
V=-mc \cdot g
\end{equation}
\begin {equation}
T=\frac{m}{4}\left(v_p \cdot v_p+v_q \cdot v_q\right)
\end{equation}
\begin{equation}
\forall i\in\{1...|q|\}, \frac{d}{dt} \left(\frac{\partial L}{\partial \dot{q_i}}\right) - \frac{\partial L}{\partial q_i}=Q
\end{equation}

%Taking the time derivative of the angles pictured above yields \begin{equation}
%\omega=[ 0, \dot{\alpha}, \dot{\beta} ]; .  \end{equation}
Note that the rotation around the $\hat{x}$ axis will be disregarded by this
model, since the rod is massless, and the end masses are point masses, rendering
a rotational inertia value of zero around this axis. 
%\begin {equation} v_p=v_{c} - l\times r \end{equation} \begin{equation}
%v_q=v_{c}+l \times r \end{equation} \begin {equation} c=[x,  y,  z]^T
%\end{equation} \begin{equation} r= \begin{bmatrix} r_x\\   r_y\\   r_z
%\end{bmatrix} = \begin{bmatrix} cos(\alpha)cos(\beta)\\
%cos(\alpha)sin(\beta)\\    sin(\alpha) \end{bmatrix} \end{equation}
This results in the following equations: \begin {equation}
\frac{d}{dt}\left(\frac{\partial L}{\partial \dot{x}}\right)-\frac{\partial
L}{\partial x}=Q \Rightarrow \ddot{x}=\frac{(F_p+F_q)}{m}\cdot \hat{x}
\end{equation} \begin {equation} \frac{d}{dt}\left(\frac{\partial L}{\partial
\dot{y}}\right)-\frac{\partial L}{\partial y}=Q \Rightarrow
\ddot{y}=\frac{(F_p+F_q)}{m}\cdot \hat{y} \end{equation} \begin {equation}
\frac{d}{dt}\left(\frac{\partial L}{\partial \dot{z}}\right)-\frac{\partial
L}{\partial z}=Q \Rightarrow \ddot{z}=\frac{(F_p+F_q)}{m}\cdot \hat{z}
\end{equation} \begin{multline} \frac{d}{dt} \left(\frac{\partial L} {\partial
\dot{\alpha}}\right)-\frac{\partial L}{\partial \alpha} \Rightarrow
\ddot{\alpha}=-\dot{\beta}^2cos(\alpha)sin(\alpha)+\\
\frac{1}{ml^2}((F_p+F_q)\times \frac{l}{2}r)\cdot \hat{y} \end{multline}
\begin{multline} \frac{d}{dt}\left(\frac{\partial L} {\partial
\dot{\beta}}\right)-\frac{\partial L}{\partial \beta} \Rightarrow
\ddot{\beta}=-\dot{\beta}\dot{\alpha}cos(\alpha)sin(\alpha)+\\
\frac{1}{ml^2}((F_p+F_q)\times \frac{l}{2}r)\cdot \hat{z} \end{multline} }

\/*All units used in these models are SI unless otherwise stated. */
\end{document}